\documentclass[pdflatex,sn-mathphys-num]{sn-jnl}

\usepackage{graphicx}%
\usepackage{multirow}%
\usepackage{amsmath,amssymb,amsfonts}%
\usepackage{amsthm}%
\usepackage{mathrsfs}%
\usepackage[title]{appendix}%
\usepackage{xcolor}%
\usepackage{textcomp}%
\usepackage{manyfoot}%
\usepackage{booktabs}%
\usepackage{algorithm}%
\usepackage{algorithmicx}%
\usepackage{algpseudocode}%
\usepackage{listings}%
\usepackage{longtable}%
\usepackage{comment}%
\usepackage{pdflscape}

\theoremstyle{thmstyleone}%
\theoremstyle{thmstyletwo}%

\theoremstyle{thmstylethree}%

\begin{document}

\title[Hierarchical Classification via Cascading Feature Elimination: Application to Human Phenotype Ontology-Aligned Facial Phenotyping (FaceMesh2HPO)]{Hierarchical Classification via Cascading Feature Elimination: Application to Human Phenotype Ontology-Aligned Facial Phenotyping (FaceMesh2HPO)}

\author*[1]{\fnm{Fabio} \sur{Hellmann}}\email{fabio.hellmann@informatik.uni-augsburg.de}

\author[2]{\fnm{Alexander} \sur{Hustinx}}\email{alexander.hustinx@uni-bonn.de}

\author[3]{\fnm{Benjamin D.} \sur{Solomon}}\email{solomonb@mail.nih.gov}

\author[]{\fnm{GestaltMatcher} \sur{Database Consortium}}

\author[2]{\fnm{Tzung-Chien} \sur{Hsieh}}\email{thsieh@uni-bonn.de}

\author[2]{\fnm{Peter} \sur{Krawitz}}\email{pkrawitz@uni-bonn.de}

\author[1]{\fnm{Elisabeth} \sur{André}}\email{andre@informatik.uni-augsburg.de}

\affil*[1]{\orgdiv{Chair for Human-centered Artificial Intelligence}, \orgname{University of Augsburg}, \orgaddress{\street{Universitätsstr. 6a}, \city{Augsburg}, \postcode{86159}, \state{Bavaria}, \country{Germany}}}

\affil[2]{\orgdiv{Institute for Genomic Statistics and Bioinformatics}, \orgname{University of Bonn}, \orgaddress{\street{Venusberg-Campus 1}, \city{Bonn}, \postcode{53127}, \state{North Rhine-Westphalia}, \country{Germany}}}

\affil[3]{\orgdiv{National Human Genome Research Institute}, \orgname{National Institutes of Health}, \orgaddress{\street{10 Center Dr}, \city{Bethesda}, \postcode{20892}, \state{MD}, \country{United States of America}}}




\abstract{
    \textbf{Purpose:} Many genetic disorders manifest with facial phenotypes, and clinicians are trained to recognize specific traits or morphological relationships to support the diagnostic process. However, current image-based methods output syndrome-level predictions in a "black-box" manner and do not directly support the structured description of facial morphology. Here, we introduce FaceMesh2HPO, a framework for classifying facial phenotypic descriptors aligned with the Human Phenotype Ontology (HPO) to support the diagnostic process.
    
    \textbf{Methods:} A panel of 124 clinicians manually annotated a subset of GestaltMatcher Database images for 10 disorders with 107 total HPO terms (59 leaf and 48 parent terms), refining the original curation. We combined these annotations with non-syndromic reference faces from UTKFace and extracted 3D facial meshes with 478 automatically detected points from 2D images. We then trained a hierarchical, cascading classification pipeline of PointNet-based models organized along the HPO tree, using dynamically parameterized architectures and iterative point elimination that prunes mesh points according to term-specific importance - also called feature elimination.
    
    \textbf{Results:} The best-performing configuration used 3D face meshes including the facial outline together with age, sex, and ethnicity metadata, a point-importance threshold of $0.01$, and soft labels of $0.05$ for negative samples. This configuration achieved a mean AUROC of $0.750\pm0.042$ across HPO models in cross-validation, with 3D outperforming 2D and both facial outline and metadata consistently improving performance. Per-node mean AUROCs ranged from $\approx0.55$ to $\approx0.89$, with parent and “compression” nodes near the root generally outperforming leaf nodes. The top five HPO models achieved AUROCs of $\approx0.89$ and F1-Scores of $\approx0.89$, whereas the weakest leaf models remained close to chance. On an external, independent test set, aggregated, normalized mean F1-score differences between test and validation varied by disorder: some unseen syndromes (e.g., Seckel and Sotos syndromes) showed small differences, whereas others (e.g., Mowat–Wilson, Nicolaides–Baraitser, Floating–Harbor, FBXW7, and White–Sutton syndromes) showed larger deviations, indicating heterogeneous generalizability across disorders and ontology levels.
    
    \textbf{Conclusion:} Geometric representations of the face via 3D meshes, combined with a hierarchical PointNet architecture and cascaded point elimination along the HPO hierarchy, enable clinically meaningful classification of facial phenotypes. The model partially transfers to unseen disorders, especially at the level of parent HPO terms, while performance for specific leaf phenotypes remains constrained by underrepresented labels. These findings highlight the need for more diverse training cohorts, improved treatment of rare terms during point elimination, and strategies that reconcile model-driven point selection with expert-defined region masks to enhance interpretability and robustness. Integrated into clinical workflows, FaceMesh2HPO could streamline the structured phenotypic description of patients and provide interpretable, ontology-linked support for experts during the diagnostic process.
}




\maketitle

\section{Introduction}\label{introduction}
Patients with genetic disorders often go through protracted diagnostic odysseys before receiving a diagnosis, which is partly due to limited physician specialist availability and  non-specialists' lack of familiarity with these conditions~\citep{bauskis_diagnostic_2022,michaels-igbokwe_standardized_2021}. Due to the heterogeneous presentations and individual rarity of genetic disorders, diagnosis can be challenging. Facial dysmorphisms manifest in many genetic disorders and play a key role in clinical genetics diagnosis~\citep{hsieh_computational_2023}. The availability of digital facial images in clinical workflows and registries enables next-generation phenotyping tools that leverage computer vision and deep learning to detect syndromes from 2D facial photographs~\citep{hsieh_gestaltmatcher_2022,hustinx_improving_2023,van_der_donk_next-generation_2019,dingemans_phenoscore_2023}. These tools offer new possibilities for supporting clinicians in the diagnostic process, potentially detecting subtle details and identifying conditions that may be difficult to recognize. Most existing methods, however, are designed to directly predict disorders~\citep{hsieh_gestaltmatcher_2022,hustinx_improving_2023,van_der_donk_next-generation_2019,dingemans_phenoscore_2023}. While effective in well-represented settings, such as more common and easily recognized pediatric conditions in European ancestry populations, this paradigm introduces important limitations~\citep{hsieh_computational_2023}. Performance typically depends on large, well-annotated datasets and may vary across ancestry and age groups, while limited interpretability constrains clinical trust and adoption~\citep{Duong2024}. More fundamentally, disorder-level classification ties predictions to a predefined set of known syndromes, restricting generalization to previously unseen conditions and offering limited insight into the phenotypic traits that drive predictions.

In clinical practice, since a single obvious syndrome is often challenging to identify, the differential diagnosis process involves systematically evaluating phenotypic traits. This process is commonly structured using the Human Phenotype Ontology (HPO)~\citep{Robinson2008}, a standardized, graph-based vocabulary that captures abnormalities across physiological and psychological traits, including detailed facial traits corresponding to variations in facial morphology~\citep{Qiao2024}. This suggests that facial analysis can be more naturally framed as the prediction of phenotypic traits rather than direct syndrome classification. Modeling facial dysmorphisms at the level of phenotypic traits enables more granular and reusable representations, supports interpretability by linking predictions to clinically meaningful traits, and allows models to capture patterns shared across disorders. Importantly, this approach aligns machine learning outputs with clinical reasoning and remains informative even when no exact syndrome match is available, prioritizing generalization and clinical utility over benchmark performance. Despite these advantages, HPO-based facial modeling poses practical challenges. Available annotations are often sparse, presence-only, and incomplete, leading to ambiguous negatives, while hierarchical dependencies among HPO terms introduce additional complexity. To address these issues, we leverage the HPO hierarchy while simplifying it by removing cross-links and explicitly accounting for label incompleteness at the phenotype level.

To date, the GestaltMatcher DataBase (GMDB)~\citep{lesmann_gestaltmatcher_2024,hsieh_gestaltmatcher_2022} represents the largest curated, FAIR-compliant dataset of facial photographs, cytogenomic and molecular diagnoses, and clinical annotations for rare genetic disorders. It serves as both a reference resource and a foundation for developing phenotyping tools. While GMDB includes HPO annotations, these are typically sparse and derived from publications or clinical reports, resulting in weakly labeled data. To systematically evaluate facial phenotype prediction, we focus on a subset of disorders with sufficient representation and well-characterized facial dysmorphisms, complemented by targeted, image-centric HPO curation to improve the quality of labels for facial traits.

To better reflect phenotypic structure, we move from pixel-based images to geometric 3D face meshes, which are more robust to pose and illumination, align naturally with anatomical regions, and facilitate interpretability at the level of facial regions and points. With our approach, the focus is on the hierarchical HPO tree structure without cross-links, where each HPO term is represented as an individual model. Given the specific focus of the HPO leafs, we further propose including information reduction into the training cycle, since not all mesh points are relevant. Therefore, we introduce a cascading training strategy in which each HPO term model uses data from its descendants and passes point masks to its children. After each model's training, a point importance analysis identifies unimportant points and eliminates them for the children - also called feature elimination.

Together, this work introduces a hierarchical, phenotype-centered framework for predicting facial HPO terms from 3D facial meshes, along with a large, clinician-curated benchmark with explicit positive and negative annotations for facially observable HPO terms. This combination of methodological and data contributions complements existing disorder classification systems by emphasizing interpretability and clinically aligned generalization.

\section{Related Work}\label{relatedwork}
In this section, we contrast syndrome (genetic condition) and phenotypic descriptors classification. We briefly discuss key concepts and how our approach differs from direct syndrome classification.

\subsection{Syndrome Classification}
The classification of syndromes using 2D images is a major research area in clinical genetics. Works like GestaltMatcher~\cite{hsieh_gestaltmatcher_2022} provide a useful classification tool for clinicians; this approach utilizes a pretrained deep Convolutional Neural Network (CNN) and finetunes it for syndrome classification. With their improved version with GestaltMatcher-Arc~\cite{hustinx_improving_2023} they used a model ensemble of iResNet and ArcFace with cross entropy loss. In addition, Islam et al.~\cite{islam_lightweight_2025} investigated smaller CNN architectures (e.g., ResNet50, VGG16, VGG19, AlexNet) and added an attention layer to these networks to improve performance. They showed that the VGG16 model performed best despite their smaller architectural build.
However, these approaches focus solely on 2D images and on the direct classification of syndromes with (deep) CNNs, which makes it difficult to explain how they achieve their results and, as a result, does not help physicians understand how specific facial traits may contribute to syndrome classification.

Another promising approach leverages 3D images, where the third dimension provides information gain. Hallgrimsson et al.~\cite{hallgrimsson_automated_2020} investigated faces of people with genetic disorders as well as healthy individuals based on facial shape differences in order to classify syndromes using parametric and machine learning methods. Another approach using 3D images was introduced by Banniser et al.~\cite{bannister_deep_2022}, in which faces of people with genetic disorders were generated alongside their healthy counterparts (so-called counterfactuals), and the syndrome class was determined using Bayes' theorem. Mahdi et al.~\cite{mahdi_multi-scale_2022} classified syndromes with 3D images but extracted facial regions (e.g., nose, mouth) from the 3D face and classified the facial region and the rest of the face separately before merging the results for the final outcome. Therefore, they used geometric encoding to reduce the complexity of the 3D data and linear discrimination analysis for classification. Overall, these methods show promising results, but the third dimension increases data complexity and, consequently, usually requires more data for models to achieve higher performance. In our approach, we use less dense 3D face images with lower complexity, thereby reducing the data required for high performance. 

In a combined version of 2D images and geometric information in the form of a face mesh, Dingemans et al.~\cite{dingemans_comparing_2022} used a CNN (VGGFace2) pipeline to process 2D images and, in parallel, compute edges from the face mesh extracted from the 2D images to feed their results into a Bayesian classifier for syndrome classification. However, they used the distance between face mesh points rather than the points themselves and, more importantly, predicted disorders rather than HPO-based phenotypes, which differs from our approach. We aim to classify facial phenotype traits, not syndromes. Dingemans et al.~\cite{dingemans_phenoscore_2023} developed PhenoScore, where they also used HPO terms as a second input, in addition to 2D images, to enhance information gain when classifying syndromes. However, it is possible that the provided HPO terms can already classify a syndrome without the need for analysis of the 2D image. In addition, clinicians would have to identify HPO terms themselves before obtaining a result about the possible underlying disorder. Therefore, we propose a direct approach to classify HPO terms, potentially improving disorder classification results. 

\subsection{Phenotypic Descriptor Classification}
An example of the kind of work our method could support is that of Robinson et al.~\cite{robinson_interpretable_2020}, which uses their LIRICAL framework to compute the probabilities of disorders based on underlying HPO terms via likelihood ratios. Clinicians can already benefit from the LIRICAL framework by identifying HPO terms themselves and providing them to the framework. Finally, the HPO term classification can also be addressed from the genotypic side, as Ullah et al.~\cite{ullah_estimating_2013} demonstrated using a bipartite network, since there are lists of known genes connected to one or more HPO terms. However, genetic and genomic testing is not universally accessible, and can be slow and expensive,  meaning that it is not always feasible for physicians to rely on such approaches.

\section{Dataset}\label{dataset}
The main dataset used in this work is the GestaltMatcher DataBase (GMDB)~\cite{lesmann_gestaltmatcher_2024}. The GMDB is a curated, FAIR-compliant dataset of clinical and other phenotypic data from individuals with genetic disorders, with a particular focus on facial dysmorphism. It contains annotated patient cases, including frontal face photographs and, when available, molecular or cytogenomic diagnoses. The dataset further includes clinical descriptive traits (HPO terms), sex, age, ancestry, pedigree information, and other relevant medical images and data, with varied availability and completeness across cases. 
GMDB serves both as a reference resource for clinicians and medical researchers and as a benchmark and training dataset for next-generation phenotyping tools and methods, such as GestaltMatcher(-Arc)~\cite{hsieh_gestaltmatcher_2022,hustinx_improving_2023}. 

\subsection{GMDB-HPO}
\label{sec:gmdb_hpo}
For this work, we focused on a subset of disorder groups that have both relatively high representation in GMDB and notable facial dysmorphism. Disorder groups include all subtypes of a disorder; hereafter, these are collectively referred to as disorders. For instance, Kabuki syndrome has subtypes 1 and 2 (caused by variants in two different genes), which are in this work both considered as Kabuki syndrome. Although subtypes may differ, they share key dysmorphisms. Based on these criteria and reflecting a practical trade-off between dataset diversity and manual annotation effort, we selected 10 disorders for downstream analyses in this initial study. After all rounds of quality control, this subset contains 1,230 images of 1,001 individuals, with a total of 16,569 present and 34,355 absent HPO-term annotations across 72 HPO terms. However, in the training process only 70 HPO terms were used, as two HPO terms (HP:0410030 - Cleft lip and HP:0004493 - Craniofacial hyperostosis) had insufficient data. Furthermore, all respective parent HPO terms were added, yielding a final set of 107 HPO terms (as described in Section~\ref{sec:preprocessing}).
Table~\ref{tab:gmdb_hpo_samples} shows the disorders and their patient and image distributions in GMDB-HPO.

\begin{table}[h!]
    \centering
    \caption{Number of unique individuals and images per disorder in GMDB-HPO.}
    \label{tab:gmdb_hpo_samples}
    \begin{tabular}{lrr}
        \toprule
        Disorders & \# Individuals & \# Images \\
        \midrule
        Cornelia de Lange syndrome (CdLS)      & 171  & 183  \\
        Kabuki syndrome (KS)                   & 128  & 146  \\
        Noonan syndrome (NS)                   & 130  & 144  \\
        Williams-Beuren syndrome (WBS)        & 128  & 133  \\
        Ogden syndrome (OGDNS)                & 64   & 122  \\
        KBG syndrome (KBGS)                   & 102  & 120  \\
        Coffin-Siris syndrome (CSS)           & 96   & 118  \\
        Angelman syndrome (AS)                & 83   & 99   \\
        Schaaf-Yang syndrome (SHFYNG)         & 35   & 85   \\
        Mucopolysaccharidoses (MPS)           & 64   & 80   \\
        \midrule
        Total                                 & 1,001 & 1,230 \\
        \bottomrule
    \end{tabular}
\end{table}

\subsubsection{Annotation Effort}
\begin{figure}
    \centering
    \includegraphics[width=1\linewidth]{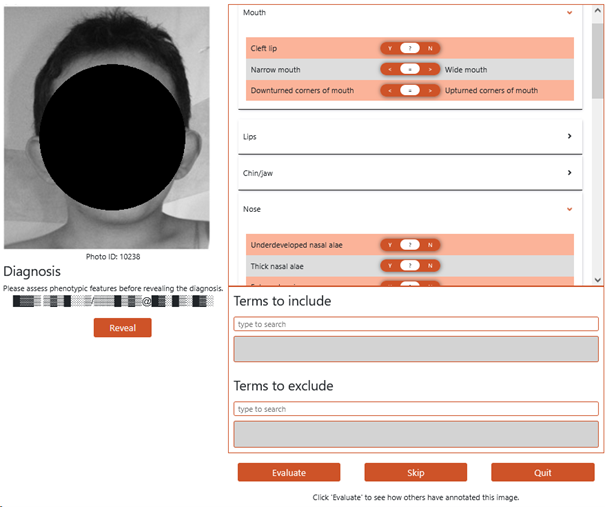}
    \caption{Face2HPO annotation tool with the patient’s face obfuscated.}
    \label{fig:annotation_tool}
\end{figure}

As GMDB is weakly labeled with respect to HPO terms (terms annotated as present are known to be present, but unannotated terms are not necessarily absent), we designed an annotation protocol and annotation tool to address this incompleteness. Initially, the OMIM and ORPHA resources for each disorder were reviewed via the HPO website, and all frequent HPO terms within the “Head and neck”, “Ear”, and “Eye” categories were collected~\cite{Hamosh2004,NguengangWakap2019,Orphadata,Robinson2008}. With the assistance of medical genetics experts, this list was further pruned to prioritize terms that are typically observable in frontal face photographs. In addition, mutually exclusive terms (e.g., “Narrow mouth” vs “Wide mouth”) were identified to streamline the annotation process. 
The annotation protocol prompts annotators to label each pre-selected term as “present”, “absent”, or “uncertain” (when visibility or interpretation is unclear), and also allows annotators to add additional relevant HPO terms not included in the pre-selected list of HPO terms. To minimize annotation bias, the patient’s disorder label is initially hidden during annotation. The annotation tool is shown in Figure~\ref{fig:annotation_tool}.

In total, 124 annotators contributed to the annotation effort. Annotators were recruited in multiple rounds through the GMDB platform. The participants were GMDB users with medical backgrounds who expressed interest in the study. Participation was voluntary, and contributors with the highest annotation activity were eligible for co-authorship, mentions, or acknowledgments in accordance with standard authorship guidelines. All annotations were performed in accordance with GMDB's data access and governance policies. The annotation effort began in January 2025; however, for the purpose of this study, a data freeze was applied in December 2025, defining the dataset used for all analyses reported here.

\subsubsection{Quality Control}
Throughout the annotation effort, a total of 1,676 frontal face photographs were assessed and annotated. A subset of photographs exhibited annotation inconsistencies, such as the absence of any term annotated as present or the absence of any term annotated as absent. We therefore applied an initial round of quality control (QC) to remove inconsistent or incomplete annotations, resulting in 1,301 images. 
Following QC, the annotations generated in this study were combined with existing HPO annotations from GMDB for the same images, incorporating previously labeled terms. A final round of QC was then applied to resolve conflicts between newly added annotations and existing ones (e.g., a term marked as present in GMDB but annotated as absent in this study). These overlaps between existing HPO terms and new HPO annotations were resolved by following the two-step merging procedure: 1) Present traits (PT) and absent traits (AT) from the GMDB were merged with their new manual present (MPT) and absent (MAT) counterparts; 2) MPTs that existed within the AT were not added to the PT, with the same applying to MAT and AT.
After this final integration and QC, the resulting dataset comprised 1,230 photographs, with a total of 72 HPO terms annotated. A flowchart with the stages of QC is shown in appendix Fig.~\ref{fig:gmdb_hpo_flowchart}, and a full list of HPO term frequencies and their corresponding mean Inter-Rater-Agreement (IRA) with standard deviation and 95\% confidence intervals across all samples were computed using Fleiss' Kappa~\citep {fleiss_measuring_1971} are shown in appendix Table~\ref{tab:hpo_frequencies_appendix}.

\subsection{Independent Test Set}
\label{sec:independent_test_set}
In addition to GMDB-HPO, we evaluated model performance on an independent test set annotated by multiple experts. This dataset was constructed prior to the start of the large-scale annotation effort described in Section~\ref{sec:gmdb_hpo}, originally with the aim of exploring the feasibility of HPO-based classification from facial images. While its size was insufficient for training a meaningful model, the dataset provides a valuable resource for evaluation.

After similar QC, the test set comprises 17 disorders, each with up to 5 patients. For each image, three medical experts independently annotated HPO terms. In contrast to the main dataset,  where each image is annotated by a single expert selected from a large pool of annotators, this setup ensures consistent multi-expert annotation across all samples, thereby reducing inter-rater variability and increasing annotation reliability. After quality control, we used a total of 75 test images, whose distribution is shown in Table~\ref{tab:independent_test_set}.

\begin{table}[h!]
    \centering
    \caption{Disorders in the independent test set, their frequencies, and whether the disorder was included during training of the models.}
    \label{tab:independent_test_set}
    \begin{tabular}{lll}
        \toprule
        Disorder & Frequency & Trained on? \\
        \midrule
        Coffin-Siris syndrome (CSS) & 12   & Yes \\
        Cornelia de Lange syndrome (CdLS) & 5    & Yes \\
        Ogden syndrome (OGDNS) & 5    & Yes \\
        KBG syndrome (KBGS) & 5  & Yes \\
        Noonan syndrome (NS) & 4    & Yes \\
        Williams-Beuren syndrome (WBS) & 4    & Yes \\
        FBXW7 syndrome (FBXW7S) & 5    & No \\
        Floating-Harbor syndrome (FLHS) & 5    & No \\
        Hyperphosphatasia with mental retardation syndrome (HPMRS) & 5 & No \\
        Mowat-Wilson syndrome (MOWS) & 5    & No \\
        Nicolaides-Baraitser syndrome (NCBRS) & 5    & No \\
        Ohdo syndrome, SBBYS-variant (SBBYSS) & 5    & No \\
        Opitz GBBB syndrome (OGBBBS) & 5    & No \\
        Sotos syndrome (SOTOS) & 4    & No \\
        White-Sutton syndrome (WHSUS) & 4    & No \\
        Susceptibility to Autism (StA) & 2    & No \\
        Seckel syndrome (SS) & 2    & No \\
        \bottomrule
    \end{tabular}
\end{table}

The annotation protocol was designed specifically for HPO classification. For each disorder, HPO terms classified as “frequent” or “very frequent” in the Human Phenotype Ontology and Online Mendelian Inheritance in Man (OMIM)~\cite{Hamosh2004} databases were preselected. Annotators were asked to assess the presence of each term in every patient image. A term was considered present if at least two out of three annotators agreed on its presence. In addition, annotators were instructed to label any non-preselected terms that were not categorized as “frequent” or “very frequent” but were deemed clearly present; such terms were included without requiring majority agreement. Though annotators were not asked to explicitly label the absence of terms, due to the majority voting, we feel that it is reasonable to assume terms were not present if they were labeled by no more than a single annotator.

Importantly, this test set includes both disorders present in the training data and disorders not observed during training, enabling evaluation in both seen and unseen disease settings. As such, it serves as a complementary evaluation resource, providing insight into model performance under reduced annotation noise, improved label consistency, and generalizability to other disorders.

\section{Method}\label{method}
\begin{figure}[h]
    \centering
    \includegraphics[width=1\linewidth]{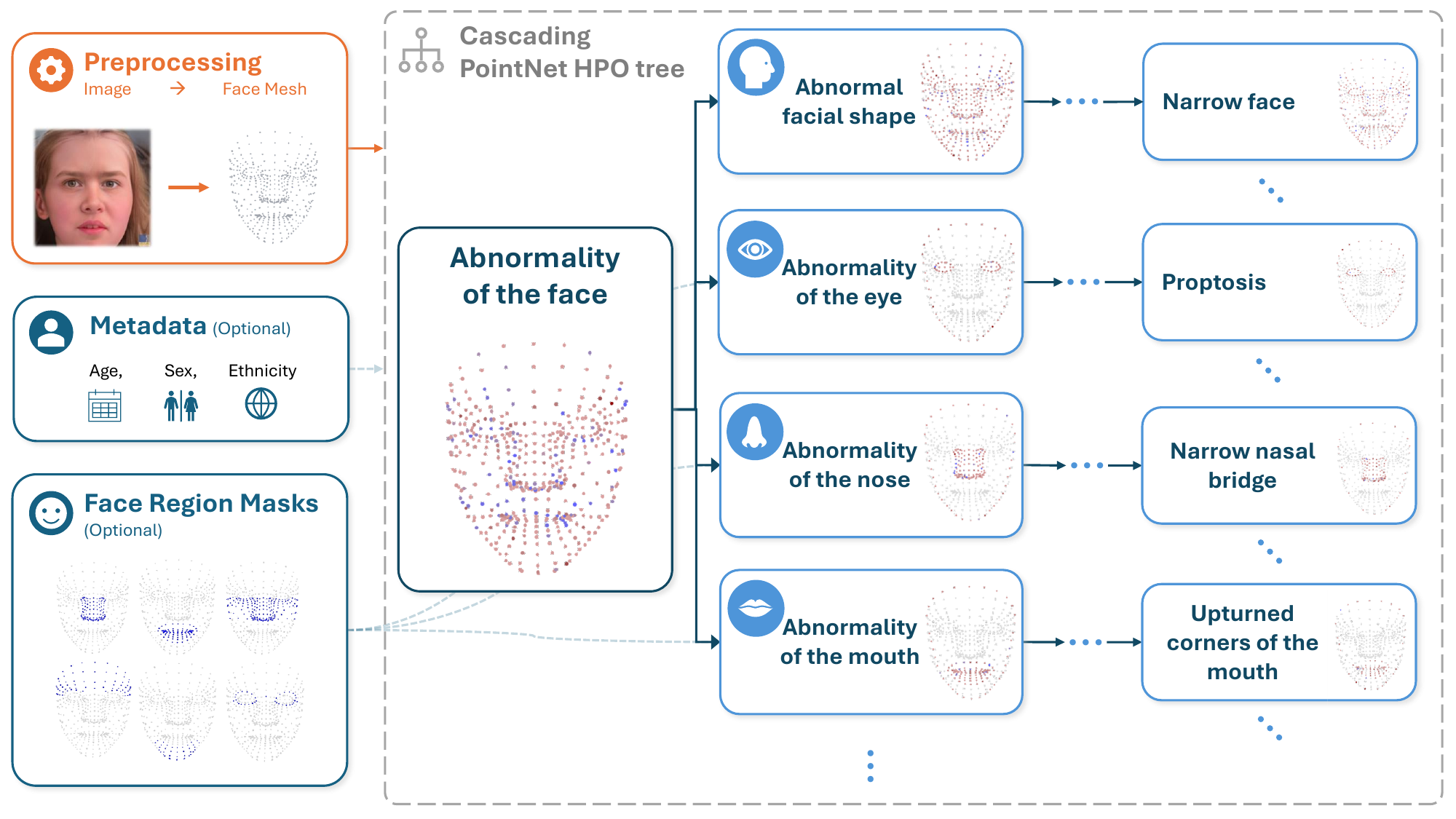}
    \caption{A schematic overview of the full FaceMesh2HPO Framework with the corresponding point importance visualizations of the face meshes for the HPO models. GestaltGAN~\cite{Kirchhoff2025} was used to synthesize the exemplary patient photograph.}
    \label{fig:framework}
\end{figure}

This section describes our methodology for cascaded hierarchical classification using feature elimination as displayed in Figure~\ref{fig:framework}. The framework includes a preprocessing step (\ref{sec:preprocessing}), optional metadata parameters (age, sex, ethnicity), face-region masks, and a cascading HPO model tree.

The HPO is not a tree structure with single connections between nodes. Cross-links between multiple nodes are possible, making it more like a graph than a tree. However, we had to cut cross-links to simplify the graph into a tree with a clear hierarchical structure, as multiple parents cannot be processed by our method. The Abnormality of the face is defined as the root node. Since the Abnormality of the eye and Abnormal eyebrow morphology are not part of the branches outgoing from Abnormality of the face under the HPO, we moved them to be children of Abnormality of the face. To replicate this tree structure, each model is defined as a single HPO term (see Figure~\ref{fig:framework}). This means that each model has a parent model and can have child models if it is not a leaf.

\subsection{Preprocessing}\label{sec:preprocessing}
The data preprocessing pipeline consists of two steps:

\textbf{1) Face Mesh Extraction:} The input data for the pipeline consists of 2D images of faces. Landmarks are extracted from these images using the Face Mesh Detector from mediapipe~\cite{kartynnik_real-time_2019} as of September 15, 2022. This detector provides 478 points distributed across the face in a semi-3D face mesh, with the third dimension approximated.

\textbf{2) HPO Table Generation:} For each image, the phenotypes are read and displayed in a table. Since the phenotypes are predominantly leaves in the HPO tree, their parents must be identified and included in the table so that each image records the corresponding phenotype, starting from the root of the HPO tree.

\subsection{Metrics}\label{metrics}
The models were evaluated using several metrics. The Area Under the Receiver Operating Characteristic Curve (AUROC)~\citep{hanley_meaning_1982} is the primary metric for assessing a model's performance. Because it reflects the $sensitivity$ and $1-specificity$ balance, it provides a clear indication of a model's performance. An AUROC of 0.9 to 1.0 is excellent, 0.8 to 0.9 is good, 0.7 to 0.8 is moderate, 0.6 to 0.7 is poor, and 0.5 to 0.6 is near chance.
In addition, the Matthews Correlation Coefficient (MCC)~\citep{matthews_comparison_1975} is used to select the best model for inference, as it provides a more fine-grained score using sensitivity, specificity, precision, and negative predictive value. Therefore, a high MCC (e.g., MCC=0.95) always corresponds to a high AUROC but not vice versa~\citep{Chicco2023}.
Furthermore, the metrics F1-Score, precision, and recall were chosen to provide a general impression of the models' performance. Additionally, the prevalence and detection prevalence are provided for better comparison.

\subsection{Model-Architecture}\label{model}
\begin{figure}
    \centering
    \includegraphics[width=1\linewidth]{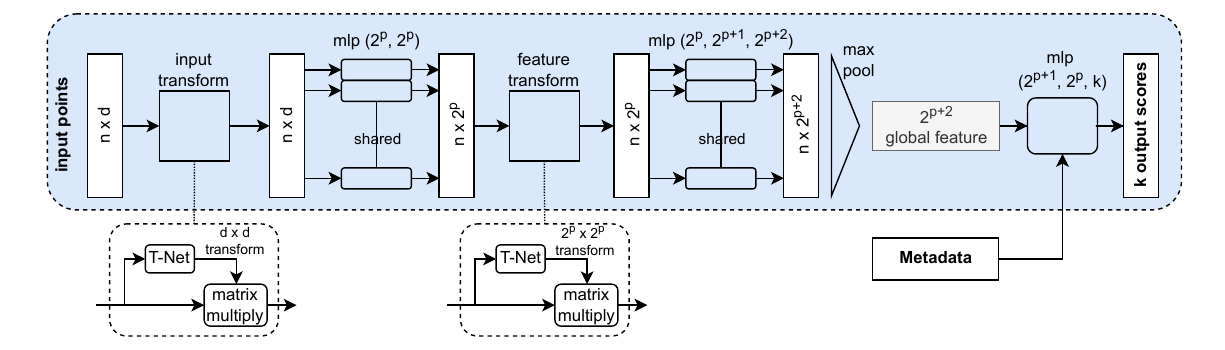}
    \caption{The modified PointNet architecture for increased flexibility.}
    \label{fig:pointnet}
\end{figure}
The architecture used for the HPO model tree is based on the PointNet~\cite{charles_pointnet_2017} architecture as shown in Figure~\ref{fig:pointnet}. The number and type of layers remain the same. The input size is based on the dimensions $d$ used. To increase flexibility when working with different point counts, we dynamically adjusted the model's parameter count by varying $p$. The reduction of the parameter count depends on the number of input points. When more than 100 input points are available, $p=6$. If the number is between 50 and 100, $p=5$, and in a range greater than 25 and 50, $p=4$. Finally, when the input point number is 25 or less, $p=3$. If the number of input points falls below a critical threshold for processing by the architecture, a zero-padded difference between the minimum and the actual number of points is added after the input transformation. To embed the metadata into the classification task, the values are converted to integer numbers: -1 if not available, or a dedicated number for each class. The metadata list is added to the global features and processed by the last Multi-Layer-Perceptron (MLP) step.
Furthermore, we added a more robust normalization to each convolutional layer, switching to group normalization when the batch size is 1, as batch normalization fails. If the batch size is greater than 1, the batch normalization is applied.

\subsection{Training Strategy}\label{training}
The tree root is trained first using all available points from the face mesh. For the input data, all samples, including itself and its children, are used recursively. That means the root model is trained on the entire dataset of affected individuals, and the unaffected samples are drawn from the UTKFace~\citep{zhang_age_2017} dataset, with a distribution as close as possible to that of our dataset in age, sex, and ethnicity. At higher levels of the tree, when unaffected samples are available in both our dataset and the UTKFace dataset, we split at most 50/50 between the two.

After a model is trained on a set of points, a point-wise feature importance is computed using the Integrated Gradients~\citep{sundararajan_axiomatic_2017} to extract importance values for each point on the face mesh. The feature importance of each fold is accumulated through all folds, and the mean is computed to gather a global feature importance of all models. Based on the global feature importance, a threshold is applied to a mask, eliminating all points that fall below it. This is represented in Fig.~\ref{fig:framework}, where the color of the points in each HPO node indicates the Integrated Gradient feature importance of each point in the training set. The more blue a point is, the higher the chance that the point is eliminated. The eliminated global feature importance mask is used in the next iteration of the model's children, limiting them to only the points from the face meshes present in the mask. This cascading reduction of points across each level of the HPO tree simplifies the classification task. For some HPO models (Abnormality of the mouth, Abnormality of the orbital region, Abnormality of the nose, Abnormality of the eye, Abnormal midface morphology, Abnormal forehead morphology, Abnormality of the periorbital region, Abnormality of the chin) masks were predefined by a genetic physician expert on a specific set of points to focus the classification on certain regions, e.g., mouth, eyes, nose, chin, forehead. Therefore, we developed an interactive website (\url{https://hcmlab.github.io/hpo-mesh-annotator/}) to enable medical professionals to easily select relevant region points. If a predefined mask is provided to the model, the predecessor mask is ignored.

In general, all models were trained using 5-Fold Stratified Cross Validation (CV) to ensure that a face, displayed across multiple images, does not appear in both the training and validation splits. A model was trained only if a point mask with at least 2 points was provided, and the sample size for the HPO term was greater than or equal to 50. The training was performed for up to 25 epochs, with a learning rate of 0.0001, automatic learning rate reduction at a plateau with a patience of 5 epochs, and early stopping with a patience of 5 epochs. The training seed was set to 42.

\section{Results}\label{result}
We conducted an ablation study to evaluate different model-tree configurations and identify the best configuration. Afterward, we analyzed the HPO models with the best configuration. Furthermore, a deeper look at the correlation between key metrics was taken to investigate whether correlations are present. Finally, the HPO models were evaluated on a test set.

\subsection{Model Selection through Ablation Study}
The ablation study embodies 72 experiments, which have been performed using a permutation of the following settings: dimension (3D and 2D), face outline (False and True), soft labels (0, 0.05, and 0.1), feature importance threshold (0.01, 0.05, and 0.1), and metadata (empty and [age, sex, ethnicity]). Since a face mesh is represented by an approximate third dimension, we tested whether the x- and y-coordinates alone would suffice without the z-coordinate. In addition, the face mesh's outermost points represent the face outline and can be used alongside points from a point mask. Due to the lack of unlabeled unaffected individuals, we tested whether soft labels (0, 0.05, and 0.1) could improve performance, as suggested by \citep{nguyen_learning_2014}. Since the unaffected individuals used are not verified, it is unclear whether they are indeed unaffected or might possibly be affected, which is also a reason for using soft labels. Furthermore, we tested different thresholds for the feature importance elimination (0.01, 0.05, and 0.1). Finally, we also experimented with using no metadata and age, sex, and ethnicity as metadata.

\begin{table}
\caption{The top-10 and bottom-5 results of the ablation study with 72 experiments. D=Dimensions, FO=Face Outline, T=Feature Importance Threshold, S=Soft Label.}
\label{tab:ablation_top10}
\centering
\begin{tabular}{cccccc}
\toprule
D & FO & Metadata & T & S & mean AUROC \\
\midrule
3 & True & age, gender, ethnicity & 0.01 & 0.05 & 0.750 $\pm$ 0.042 \\
3 & True & age, gender, ethnicity & 0.01 & 0.00 & 0.749 $\pm$ 0.044 \\
3 & True & age, gender, ethnicity & 0.05 & 0.10 & 0.747 $\pm$ 0.044 \\
3 & True & age, gender, ethnicity & 0.01 & 0.10 & 0.747 $\pm$ 0.043 \\
3 & True & age, gender, ethnicity & 0.05 & 0.05 & 0.741 $\pm$ 0.043 \\
3 & False & age, gender, ethnicity & 0.01 & 0.00 & 0.740 $\pm$ 0.045 \\
3 & False & age, gender, ethnicity & 0.01 & 0.05 & 0.740 $\pm$ 0.041 \\
2 & True & age, gender, ethnicity & 0.01 & 0.10 & 0.739 $\pm$ 0.043 \\
3 & False & age, gender, ethnicity & 0.01 & 0.10 & 0.739 $\pm$ 0.043 \\
3 & True & age, gender, ethnicity & 0.05 & 0.00 & 0.738 $\pm$ 0.046 \\
... & ... & ... & ... & ... & ... \\
2 & False & N/A & 0.10 & 0.10 & 0.653 $\pm$ 0.055 \\
2 & False & N/A & 0.10 & 0.05 & 0.651 $\pm$ 0.055 \\
2 & True & N/A & 0.10 & 0.00 & 0.648 $\pm$ 0.051 \\
2 & True & N/A & 0.10 & 0.10 & 0.646 $\pm$ 0.050 \\
2 & True & N/A & 0.10 & 0.05 & 0.644 $\pm$ 0.053 \\
\bottomrule
\end{tabular}
\end{table}

The results of the top-10 and bottom-5 experiments from the ablation study are listed in Table~\ref{tab:ablation_top10}. The best experiment with 3D-Face Meshes, the face outline, metadata, a threshold of 0.01, and no soft label performed with an AUROC score of $0.749\pm0.044$. The top-4 experiments all used 3D-Face Meshes, the face outline, and metadata. The changes in performance are from the used soft label, where 0.05 performed best, with 0.001 performance gain over 0.00. The feature elimination threshold of 0.01 outperformed 0.05 by $\approx0.009$ and 0.1 by $\approx0.025$. The 2D-Face Mesh experiment lagged the same configured 3D experiment by 0.007. The 3D experiment without using the face outline performed $\approx0.013$ worse than the same experiment with the face outline. A higher feature importance threshold led to lower performance, as many points were eliminated in earlier stages, leaving only a few, or even none, points for leaf-HPO terms to be trained on and, therefore, represented, resulting in zero AUROC. The full result table is available in appendix Table~\ref{tab:ablation_study_appendix}.

\subsection{HPO Delineation Performance Analysis}

\setlength{\tabcolsep}{1pt}
\begin{table}
\caption{An overview of the HPO models of the best model configuration, with a selection of the models with their top-5 and bottom-5 performing leaf HPO models and their mean scores over the 5-Fold Cross Validation. The samples (N) are 50\% affected and 50\% unaffected. The detection prevalence (Det. P) shows the model's perception of prevalence.}
\label{tab:best_model_hpos}
\centering
\begin{tabular}{rcccccc}
\toprule
 & AUROC & F1-Score & Precision & Recall & N & Det. P \\
\midrule
\textbf{Short nose} & 0.80$\pm$0.05 & 0.81$\pm$0.04 & 0.77$\pm$0.01 & 0.84$\pm$0.08 & 814 & 0.54 \\
\textbf{Wide nasal bridge} & 0.80$\pm$0.03 & 0.81$\pm$0.05 & 0.77$\pm$0.03 & 0.85$\pm$0.08 & 800 & 0.55 \\
\textbf{Hypertelorism} & 0.79$\pm$0.04 & 0.80$\pm$0.06 & 0.77$\pm$0.05 & 0.83$\pm$0.07 & 762 & 0.54 \\
\textbf{Downturned corners of mouth} & 0.78$\pm$0.03 & 0.78$\pm$0.03 & 0.81$\pm$0.08 & 0.75$\pm$0.05 & 512 & 0.47 \\
\textbf{Thin upper lip vermilion} & 0.76$\pm$0.02 & 0.78$\pm$0.03 & 0.72$\pm$0.04 & 0.85$\pm$0.05 & 1116 & 0.59 \\
... & ... & ... & ... & ... & ... & ... \\
\textbf{Long nose} & 0.60$\pm$0.05 & 0.62$\pm$0.04 & 0.59$\pm$0.04 & 0.67$\pm$0.08 & 230 & 0.57 \\
\textbf{Synophrys} & 0.59$\pm$0.03 & 0.58$\pm$0.05 & 0.60$\pm$0.05 & 0.59$\pm$0.11 & 416 & 0.49 \\
\textbf{Microphthalmia} & 0.58$\pm$0.11 & 0.42$\pm$0.31 & 0.58$\pm$0.39 & 0.42$\pm$0.33 & 66 & 0.36 \\
\textbf{Elfin facies} & 0.55$\pm$0.07 & 0.28$\pm$0.32 & 0.43$\pm$0.43 & 0.29$\pm$0.38 & 216 & 0.24 \\
\textbf{Blepharophimosis} & 0.55$\pm$0.03 & 0.47$\pm$0.20 & 0.66$\pm$0.20 & 0.47$\pm$0.27 & 152 & 0.42 \\
\bottomrule
\end{tabular}
\end{table}

\setlength{\tabcolsep}{6pt}

In Table~\ref{tab:best_model_hpos}, the performance for the top-5 and bottom-5 performing leaf HPO models is depicted. The five best HPO term models (Short nose, Wide nasal bridge, Hypertelorism, Downturned corners of mouth, and Thin upper lip vermilion) achieved AUROCs ranging from $\approx0.76$ to $\approx0.80$ and F1-Scores ranging from $\approx0.78$ to $\approx0.81$.
On the other end of the table, in the bottom-5 models, the long nose performed best with an AUROC of $\approx0.60$ and an F1-Score of $\approx0.62$. Synophrys lagged behind with an AUROC of $\approx0.59$ and F1-Score of $\approx0.58$. Microphthalmia performed third last with an AUROC of $\approx0.58$ and F1-Score of $\approx0.42$. Furthermore, Elfin facies performed second-to-last, with an AUROC of $\approx0.55$ and an F1-Score of $\approx0.28$. Finally, the worst-performing model is for Blepharophimosis, with an AUROC of $\approx0.55$ and an F1-Score of $\approx0.47$. Overall, 107 HPO models were trained.
In addition, we examined the prevalence of the training and validation sets for each CV fold and aggregated the results. Given an exact 50\% split between affected and unaffected classes, the prevalence is always 0.5. However, the detection prevalence shows a different result as shown in Table~\ref{tab:best_model_hpos}. The detection prevalence ranges from $\approx0.23$ (Elfin facies) to $\approx0.74$ (Thick vermilion border). The full table is in the appendix Table~\ref{tab:best_model_hpos_appendix}.

\begin{figure}[h]
    \centering
    \includegraphics[width=1\linewidth]{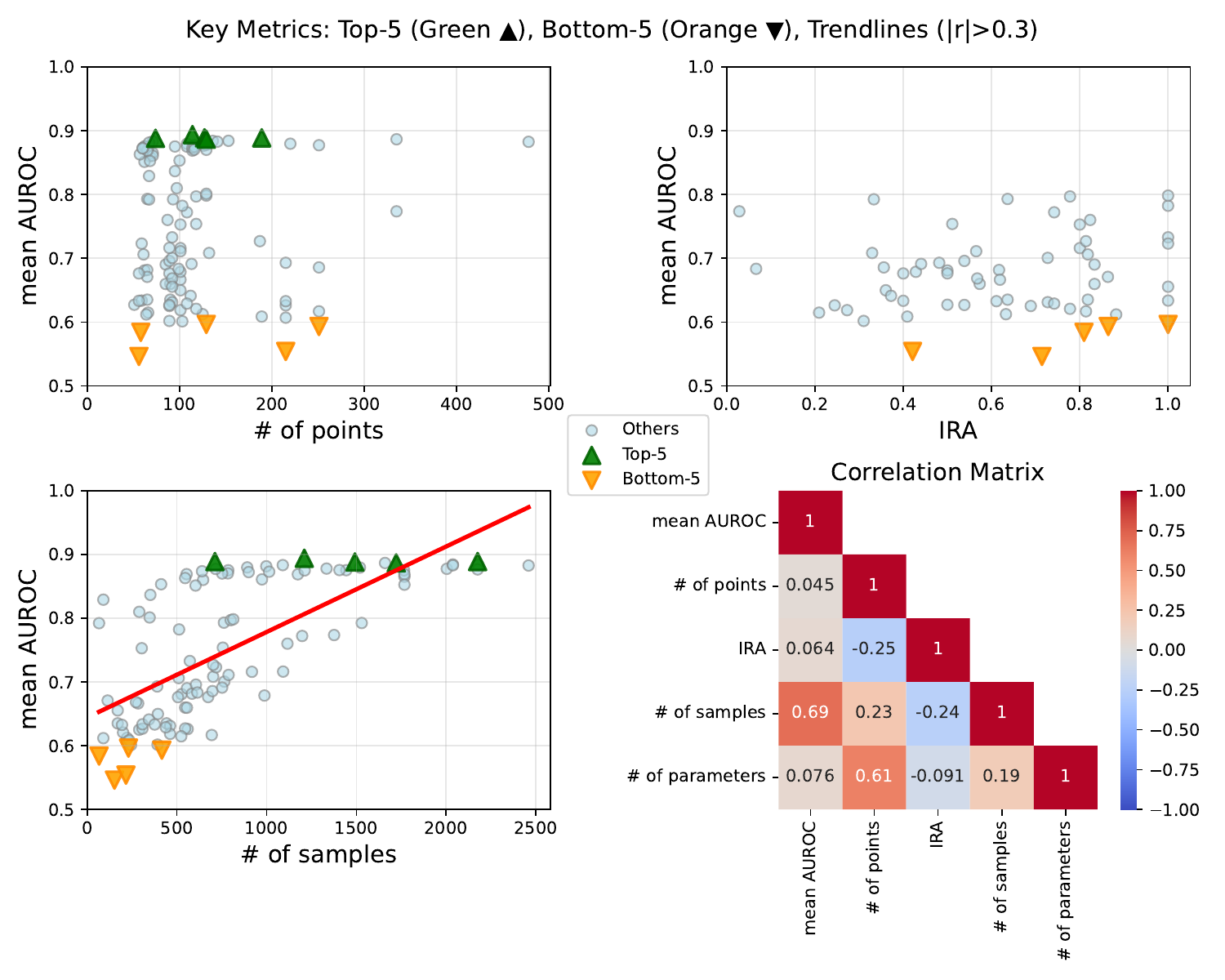}
    \caption{The correlation between key metrics: mean AUROC, number of input points, inter-rater agreement (IRA), parameter number, number of support samples.}
    \label{fig:metric_correlations}
\end{figure}

Furthermore, the Fig.~\ref{fig:metric_correlations} shows the correlation between key metrics of all HPO models with the top-5 performing models in green and bottom-5 performing models in orange. A Pearson correlation is found between the number of samples and the mean AUROC. The fewer the number of training samples, the lower the mean AUROC; the more samples, the higher the mean AUROC. No correlation was found between the number of input points and the mean AUROC. Furthermore, the Inter-Rater-Agreement (IRA) across all samples also showed no correlation with the mean AUROC. The Pearson correlation between input points and parameter count, shown in the correlation matrix with a score of $0.61$, is by design and is not coincidental.

\begin{figure}[h]
    \centering
    \includegraphics[width=0.75\linewidth]{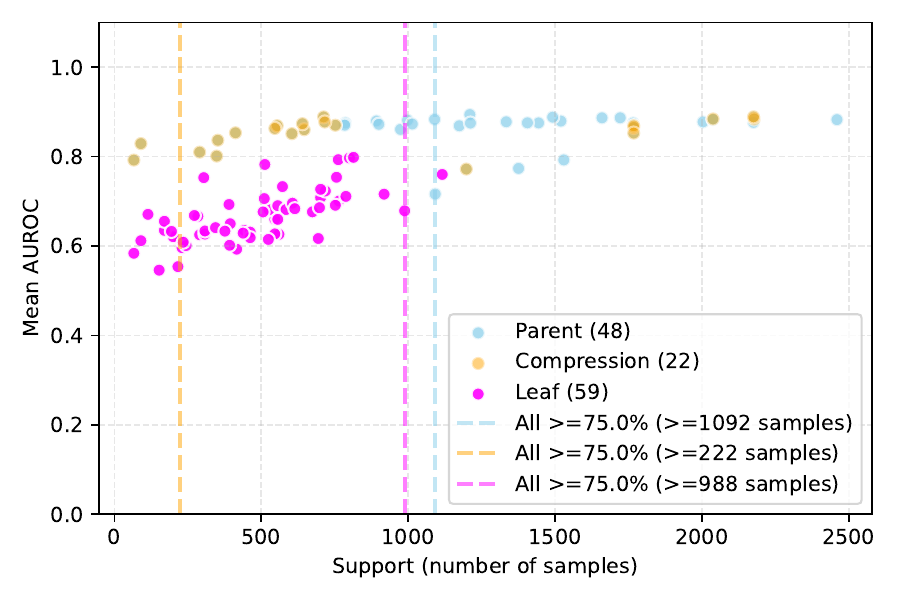}
    \caption{An in-depth analysis of the correlation of the number of samples and mean AUROC.}
    \label{fig:corr_samples_auroc}
\end{figure}
When more closely examining the correlation between the number of samples and mean AUROC (Fig.~\ref{fig:corr_samples_auroc}), the differentiation between parent and compression models and leaf models becomes clearer. A compression model has exactly one child and, as such, only compresses the dimension/number of points further down. In Fig.~\ref{fig:corr_samples_auroc}, one can see that the leaf models need at least 988 samples to achieve a mean AUROC performance of greater than or equal to $0.75$. Comparing parent and compression nodes, where the compression nodes are a subset of the parent nodes, the parent nodes need at least 1092 samples, while compression nodes need only 222 samples to achieve a mean AUROC of $0.75$ or greater.

\subsection{Evaluation of Generalizability}
We further assessed the generalizability of our approach using the independent multi-expert–annotated test set described in Section~\ref{sec:gmdb_hpo} and further detailed in the appendix. We evaluated model performance separately for disorders seen during training and those not observed during training.

\begin{figure}[h]
    \centering
    \includegraphics[width=1\linewidth]{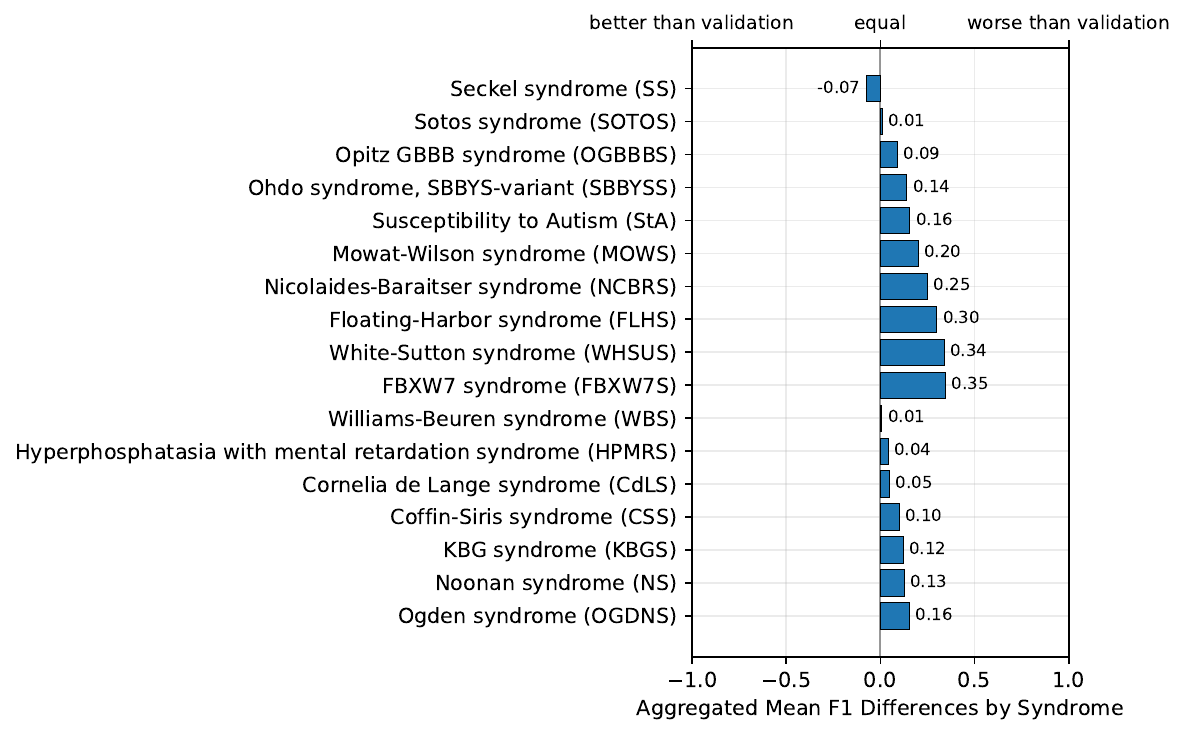}
    \caption{An overview of the aggregated mean F1-Score performance differences (test set vs. validation set) of the HPO models with a prevalence greater than zero. The difference value is normalized in the range of -1 to 1, according to the amount of available HPO term per disorder. SS to WHSUS are unseed disorders, while WBS to KBGS are seen disorders.}
    \label{fig:test_vs_val_all}
\end{figure}

Across the independent test set, the aggregated, normalized mean F1-score differences between test and validation vary by disorder, as summarized in Figure~\ref{fig:test_vs_val_all}. For unseen disorders, Seckel syndrome (SS) is the only one outperforming the validation performance of the relevant HPO terms. Sotos syndrome (SOTOS) shows small positive differences of 0.01, while Opitz GBBB syndrome (OGBBBS), Ohdo syndrome, SBBYS-variant (SBBYSS), and Susceptibility to Autism (StA) exhibit moderate differences ranging from 0.09 to 0.16. Larger differences are observed for Mowat-Wilson syndrome (MOWS), Nicolaides-Baraitser syndrome (NCBRS), Floating-Harbor syndrome (FLHS), and White-Sutton syndrome (WHSUS), with values ranging from 0.20 to 0.34. FBXW7 syndrome (FBXW7S) shows the largest observed difference of 0.35. Among disorders included during training, Williams-Beuren syndrome (WBS) yields the smallest aggregated difference of 0.01, whereas Hyperphosphatasia with mental retardation syndrome (HPMRS) and Cornelia de Lange syndrome (CdLS) show small positive differences of 0.04 and 0.05. Coffin-Siris syndrome (CSS), KBG syndrome (KBGS), Noonan syndrome (NS), and Ogden syndrome (OGDNS) exhibit larger differences of 0.10-0.16.

The two tables in the report per‑phenotype mean F1-Score differences between the independent test set and cross‑validation for all HPO terms with non‑zero prevalence, separately for unseen (appendix Table~\ref{tab:test_vs_val_unseen_appendix}) and seen disorders (appendix Table~\ref{tab:test_vs_val_seen_appendix}). For unseen disorders, the table lists how each HPO model’s F1 changes for StA, FBXW7S, FLHS, MOWS, NCBRS, SBBYSS, OGBBBS, SOTOS, SS, and WHSUS, with negative values indicating better performance on the test set and positive values indicating higher validation performance. The corresponding table for seen disorders provides the same F1 differences for CSS, CdLS, HPMRS, KBGS, NS, OGNDS, and WBS, again highlighting for each HPO term whether performance is higher in validation or on the independent test cohort.

The detailed metrics for the non-overlapping test set disorders are listed in the appendix Tables for StA~\ref{tab:pred_result_StA}, FBXW7S~\ref{tab:pred_result_FBXW7S}, FLHS~\ref{tab:pred_result_FLHS}, MOWS~\ref{tab:pred_result_MOWS}, NCBRS~\ref{tab:pred_result_NCBRS}, SBBYSS~\ref{tab:pred_result_SBBYSS}, OGBBBS~\ref{tab:pred_result_OGBBBS}, SOTOS~\ref{tab:pred_result_SOTOS}, SS~\ref{tab:pred_result_SS}, and WHSUS~\ref{tab:pred_result_WHSUS} and the disorders overlapping with the training/validation set are in the appendix Tables for CSS~\ref{tab:pred_result_CSS}, CdLS~\ref{tab:pred_result_CdLS}, HPMRS~\ref{tab:pred_result_HPMRS}, KBGS~\ref{tab:pred_result_KBGS}, NS~\ref{tab:pred_result_NS}, OGDNS~\ref{tab:pred_result_OGDNS}, and WBS~\ref{tab:pred_result_WBS}.

\subsection{FaceMesh2HPO Web Tool}

\begin{figure}[h]
    \centering
    \includegraphics[width=1\linewidth]{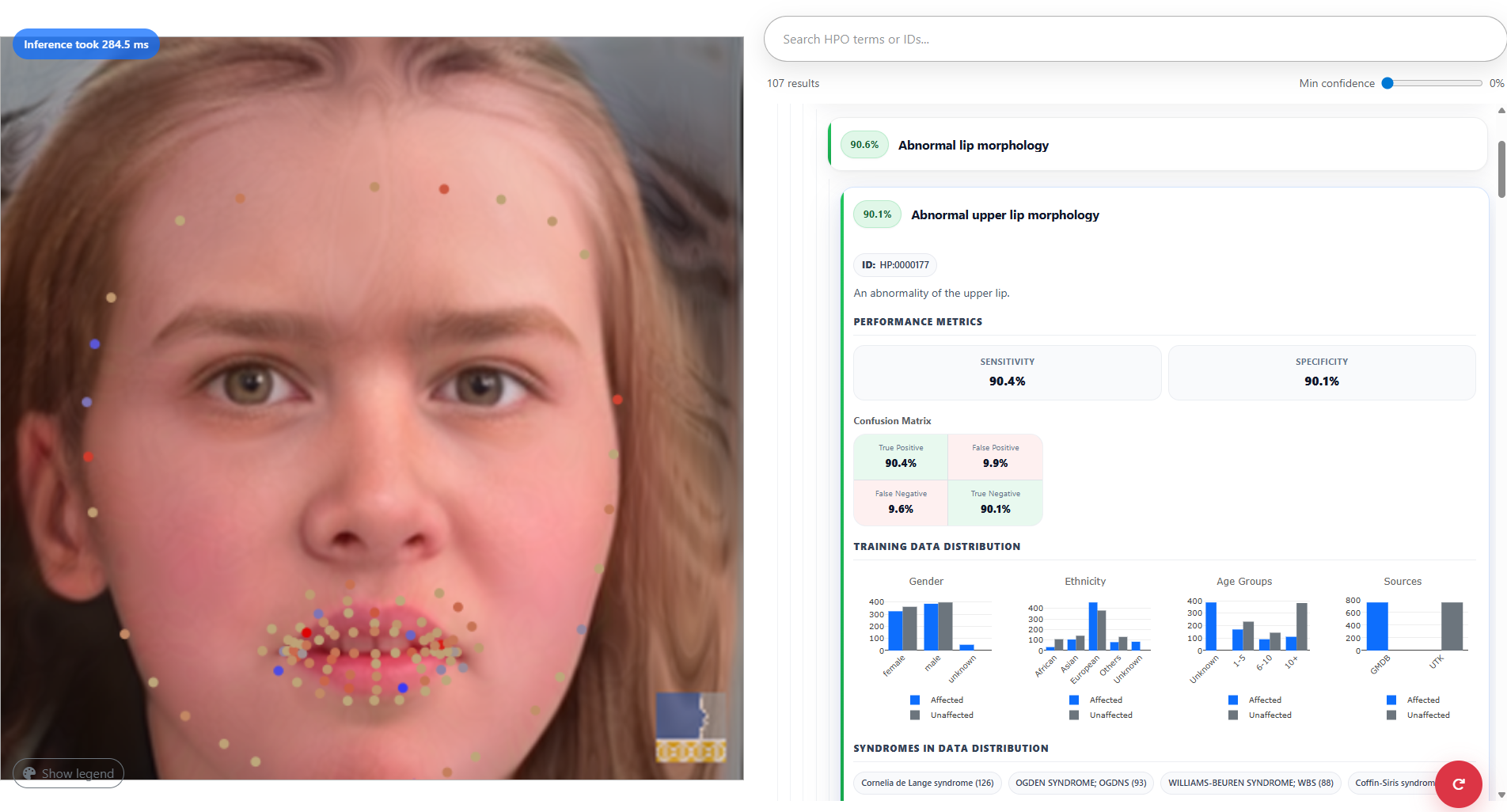}
    \caption{The detailed view of a model's summary in performance, training data, and feature importance in the FaceMesh2HPO web application. GestaltGAN~\citep{Kirchhoff2025} was used to synthesize the exemplary patient photograph.}
    \label{fig:webapp}
\end{figure}

To demonstrate the practical applicability of the FaceMesh2HPO framework, we developed a web application (\url{https://hcmlab.github.io/FaceMesh2HPO/}) that enables physicians to perform facial phenotype analyses directly on local devices (Fig.~\ref{fig:webapp}). The application integrates the best-performing model from each fold, selecting the model with the highest Matthews Correlation Coefficient~\citep{matthews_comparison_1975}. Each selected model is subsequently calibrated using either beta calibration~\citep{pmlr-v54-kull17a} (for datasets with $\ge 50$ samples) or temperature scaling~\citep{balanya_adaptive_2024} (for datasets with $< 50$ samples). The final calibration method is chosen according to the lowest Brier score~\citep{brier_verification_1950}. If the original model already achieves the lowest Brier score, no additional calibration is applied.

Upon first use, the trained models are downloaded to the local device. This one-time process may take several minutes, depending on network bandwidth, but it ensures that all subsequent analyses are performed locally without uploading patient images to external servers. Once the download is complete, the application is ready for use.

Physicians can then upload a patient image for analysis and optionally provide demographic information (age, sex, ethnicity) to refine predictions. The average total inference time, including face mesh extraction and evaluation of all 107 models, is 624.56 ms.

The results are presented in two main sections. First, the patient image is displayed alongside the extracted face mesh. Second, a hierarchical list summarizes the outputs of all models. In the default view, each entry includes the corresponding HPO term and its classification confidence, with higher values indicating greater certainty that the phenotype is present. Selecting an HPO term reveals additional details, including model performance metrics (e.g., sensitivity and specificity), dataset characteristics (e.g., age, sex, ethnicity, and geographic distribution), and the disorders used during training (Fig.~\ref{fig:webapp}, right). In the image view, classification-relevant regions are highlighted using color to indicate global feature importance (Fig.~\ref{fig:webapp}, left). Together, these features provide a transparent and interpretable overview for clinical use.

To facilitate navigation, the application includes a search function for HPO IDs and names, allowing users to quickly locate specific terms. Additionally, a confidence threshold slider enables users to filter results by setting a minimum confidence level, hiding all predictions below the selected threshold.

\section{Discussion}\label{discussion}
This work introduces FaceMesh2HPO, a hierarchical, HPO‑aligned framework for classifying facial phenotypes from 3D face meshes using a cascading training strategy with feature elimination. By assigning a dedicated classifier to each HPO term and organizing these along the HPO tree, the method learns shared structure across related phenotypes, generates predictions at multiple levels of granularity, and outputs labels that match the clinical vocabulary used in diagnostic reasoning. This structure yields rich phenotype profiles that remain informative even when the underlying disorder is unknown, and can be directly combined with tools such as LIRICAL~\citep{robinson_interpretable_2020} as compositional evidence rather than a single black‑box score.

The ablation study showed that performance is best when using 3D face meshes including the facial outline, together with age, sex, and ethnicity metadata, a feature‑elimination threshold of $0.01$, and soft negative labels of $0.05$. The modest gap between 3D and 2D meshes indicates that the approximated depth contributes only a limited additional signal, suggesting that true 3D imaging could unlock larger gains. The face outline and metadata systematically improved results, consistent with their role in providing global proportionality and demographic context for local dysmorphisms. Lower feature‑importance thresholds preserved more points for downstream nodes and improved performance, while soft labels helped mitigate uncertainty in the negative class, where apparently unaffected controls may still exhibit subtle dysmorphic traits.

Performance analyses highlight a strong dependence on ontology level and data support. Parent and compression nodes consistently outperform leaves, benefiting from larger effective sample sizes and more generic patterns. In contrast, rare leaf terms - especially those with limited geometric representation in the mesh (e.g., eyebrow or eyelash traits) often show AUROCs near chance and unstable detection prevalence, reflecting sample scarcity, the compounded effect of upstream feature pruning, and the inaccuracy of detected face meshes on dysmorphic faces as investiaged by Hellmann et al.~\citep{hellmann_ganonymization_2025} and showed that facial traits such as \emph{Big Nose} and \emph{Big Lips} seem to be not well captured by the face mesh detector. The fixed 50/50 class balancing during training, combined with noisy negatives, further contributes to miscalibration, as evidenced by the wide spread of detection prevalences across terms despite identical training prevalence. In practice, this means the framework currently offers more reliable information at coarser ontology levels and for relatively common traits than for highly specific, sparsely annotated leaves. Furthermore, we believed that the IRA would influence the models' performance: the lower the IRA score, the lower the model's performance, and vice versa. Surprisingly, the results painted a different picture, showing no correlation between the models' performance and the IRA.

The independent, multi‑expert test set clarifies how these patterns generalize. Per‑disorder analyses show that many parent and intermediate terms maintain F1-Score close to or above validation, while leaf‑level performance is more variable. These findings support the central design choice of modeling reusable HPO‑level phenotypes rather than syndromes, but also underscore that the method should be viewed as providing a phenotypic scaffold whose reliability depends on both ontology depth and disorder family, rather than a uniformly accurate detector of all rare traits.

Despite these limitations, the approach is scalable and amenable to extension. Additional HPO terms can be integrated as more curated data become available, and the mesh‑based representation keeps per‑term data needs manageable. The cascading feature elimination already reduces computational costs and focuses models on salient regions, while expert‑designed regional masks help anchor point selection in clinically meaningful areas and partially compensate for the opacity of purely data‑driven pruning.

Future work should prioritize the qualitative validation of the FaceMesh2HPO tool in a study with clinicians. Furthermore, continuing to improve support for rare terms, refining calibration under imbalanced and noisy labels, and systematically assessing performance and bias across age, sex, and ancestry. Finally, the face mesh detection must be refined for faces with dysmorphisms to receive a more accurate representation.

\section{Conclusion}\label{conclusion}
In this work, we addressed the automatic classification of facial phenotypes by utilizing Mediapipe's 3D face meshes and a dynamic PointNet architecture to handle different input sizes. We introduced hierarchical training with a feature elimination approach to reduce task complexity and simplify model classification. Therefore, the models were trained in a tree-like structure, passing a point mask to their child models. This process was done in a cascading manner. In total, 107 models were trained to classify facial phenotypes on different levels of the human phenotype ontology.

The approach demonstrates that facial phenotype classification on a geometric level, based solely on face meshes, is possible. Using 3D face meshes, the face outline, the person's age, sex, and ethnicity, a feature elimination threshold of $0.01$, and soft labels of $0.05$, the ablation study showed that these settings yielded the best overall results, with a mean AUROC of $0.750\pm0.042$. The mean AUROC across the 5-Fold CV for each HPO model ranged from $\approx0.55$ to $\approx0.89$. The better-performing HPO models are usually those closer to the root model, which is correlated with the available training sample size, as there are typically more samples for models closer to the root node. When applied to other faces with unseen disorders, the method appears reliable. However, results vary slightly per disorder. Finally, the FaceMesh2HPO tool bridges these research results to a real-world application for clinicians to use in their daily practice. The additional information provided for each model will help geneticists and other clinicians to understand and trust these and related applications.

\backmatter

%

\bmhead{Acknowledgements}
We would like to thank the GestaltMatcher Database Consortium for contributing in the curation of the HPO annotations for the GMDB-HPO dataset. The contributors with significant contributions are sorted in descending order of their annotation contributions: Ibrahim AbdElrazek, Christine Fauth, Himanshu Goel, Hellen Lesmann, Tim Ripperger, Tinatin Tkemaladze, Sofia Douzgou Houge, Ceren Alavanda, Martin Zenker, Vasilica Plaiasu, Arjan Bouman, Dorota Wicher, and Shahida Moosa.
This research was supported in part by the Intramural Research Program of the National Institutes of Health (NIH). The contributions of the NIH author(s) are considered Works of the United States Government. The findings and conclusions presented in this paper are those of the author(s) and do not necessarily reflect the views of the NIH or the U.S. Department of Health and Human Services.
As a nonnative English-speaking research team, we have limited the use of generative artificial intelligence tools to polish and enhance our English-language writing, ensuring our research findings can be disseminated globally.

\section*{Declarations}

\begin{itemize}
\item Funding: This research was supported in part by the Intramural Research Program of the National Institutes of Health (NIH) (salary for Benjamin D. Solomon). 
\item Conflict of interest/Competing interests: Not applicable.
\item Ethics approval and consent to participate: Not applicable.
\item Consent for publication: Not applicable.
\item Data availability: The curated GMDB-HPO subset is derived from the GMDB dataset (v1.1.0), which is available to qualified researchers through a controlled-access process for the development and evaluation of next-generation phenotyping methods. Researchers seeking access must explicitly state that they are requesting the GMDB-HPO dataset, obtain appropriate ethics approval, submit a research proposal via \url{https://db.gestaltmatcher.org/} or by email to info@gestaltmatcher.org, and sign a data-use agreement ensuring compliance with applicable data protection legislation, including the General Data Protection Regulation (GDPR). All requests are reviewed by the GMDB Advisory Board in accordance with legal and ethical standards. Due to the inclusion of sensitive clinical data and identifiable facial images of individuals with rare genetic disorders, the dataset is not publicly available.
\item Materials availability: Not applicable.
\item Code availability: The code will be made publicly available on GitHub (\url{https://github.com/hcmlab/FaceMesh2HPO}) upon publication.
\item Author contribution: F.H. developed the methodology, performed the experiments, and wrote the manuscript. A.H. curated the GMDB-HPO dataset and supported F.H. in designing the experiments and refining the manuscript. B.S. created the point masks for specific HPO terms to focus the model's training and revised the manuscript. The GestaltMatcher Database Consortium created the annotations for the GMDB-HPO dataset. P.K. provided helpful knowledge in the field and revised the manuscript. E.A. and T.H. revised the manuscript.
\end{itemize}

\begin{appendices}

\section{Dataset}\label{appendix:dataset}
\begin{figure}[h]
    \centering
    \includegraphics[width=0.5\linewidth]{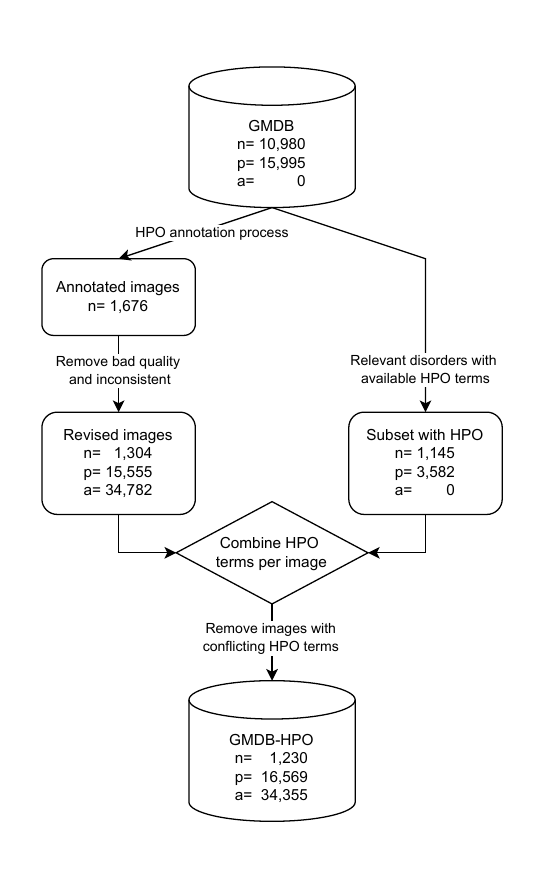}
    \caption{The flowchart represents the curation of the GMDB-HPO dataset. The number of samples (n) contains a certain amount of present phenotypes (p) and absent phenotypes (a).}
    \label{fig:gmdb_hpo_flowchart}
\end{figure}

\setlength{\tabcolsep}{3pt}
\begin{longtable}{rccr}
\caption{The frequencies of all annotated HPO terms and their corresponding Inter-Rater-Agreement (IRA) using the mean Fleiss' Kappa score for each HPO term over all samples.} \label{tab:hpo_frequencies_appendix} \\
\toprule
HPO Name & HPO ID & IRA & Samples \\
\midrule
\endfirsthead
\caption[]{The frequencies of all annotated HPO terms and their corresponding Inter-Rater-Agreement (IRA) using the mean Fleiss' Kappa score for each HPO term over all samples.} \\
\toprule
HPO Name & HPO ID & IRA & Samples \\
\midrule
\endhead
\midrule
\multicolumn{4}{r}{Continued on next page} \\
\midrule
\endfoot
\bottomrule
\endlastfoot
Abnormality of the mouth & HP:0000153 & nan $\pm$ nan & 1089 \\
Wide mouth & HP:0000154 & 0.80 $\pm$ 0.61 [-0.45, 2.05] & 459 \\
Abnormal lip morphology & HP:0000159 & nan $\pm$ nan & 1002 \\
Narrow mouth & HP:0000160 & 0.80 $\pm$ 0.61 [-0.45, 2.05] & 152 \\
Abnormal oral cavity morphology & HP:0000163 & nan $\pm$ nan & 1088 \\
Abnormal upper lip morphology & HP:0000177 & nan $\pm$ nan & 760 \\
Abnormal lower lip morphology & HP:0000178 & nan $\pm$ nan & 393 \\
Thick lower lip vermilion & HP:0000179 & 0.82 $\pm$ 0.58 [-0.37, 2.01] & 221 \\
\end{longtable}

\setlength{\tabcolsep}{6pt}

\section{Ablation Study}\label{secA1}
\begin{longtable}{ccccccc}
\caption{The results of all experiments from the ablation study. D: Dimension, FO: Face Outline, T: Feature Importance Threshold, S: Softlabel} \label{tab:ablation_study_appendix} \\
\toprule
D & FO & Metadata & T & S & mean AUROC \\
\midrule
\endfirsthead
\caption[]{The results of all experiments from the ablation study. D: Dimension, FO: Face Outline, T: Feature Importance Threshold, S: Softlabel} \\
\toprule
D & FO & Metadata & T & S & mean AUROC \\
\midrule
\endhead
\midrule
\multicolumn{6}{r}{Continued on next page} \\
\midrule
\endfoot
\bottomrule
\endlastfoot
3 & True & age, gender, ethnicity & 0.01 & 0.05 & 0.750 $\pm$ 0.042 \\
3 & True & age, gender, ethnicity & 0.01 & 0.00 & 0.749 $\pm$ 0.044 \\
3 & True & age, gender, ethnicity & 0.05 & 0.10 & 0.747 $\pm$ 0.044 \\
3 & True & age, gender, ethnicity & 0.01 & 0.10 & 0.747 $\pm$ 0.043 \\
3 & True & age, gender, ethnicity & 0.05 & 0.05 & 0.741 $\pm$ 0.043 \\
3 & False & age, gender, ethnicity & 0.01 & 0.00 & 0.740 $\pm$ 0.045 \\
3 & False & age, gender, ethnicity & 0.01 & 0.05 & 0.740 $\pm$ 0.041 \\
2 & True & age, gender, ethnicity & 0.01 & 0.10 & 0.739 $\pm$ 0.043 \\
\end{longtable}

\begin{landscape}
\section{HPO Delineation Performance Analysis}\label{secA2}
\begin{longtable}{rcccccc}
\caption{The results of all HPO models with the best configuration. The HPO terms in bold letters are leafs of the human phenotype ontology. N: Samples, Prev: Prevalence, Det. P: Detection Prevalence} \label{tab:best_model_hpos_appendix} \\
\toprule
 & AUROC & F1-Score & Precision & Recall & N & Det. P \\
\midrule
\endfirsthead
\caption[]{The results of all HPO models with the best configuration. The HPO terms in bold letters are leafs of the human phenotype ontology. N: Samples, Prev: Prevalence, Det. P: Detection Prevalence} \\
\toprule
 & AUROC & F1-Score & Precision & Recall & N & Det. P \\
\midrule
\endhead
\midrule
\multicolumn{7}{r}{Continued on next page} \\
\midrule
\endfoot
\bottomrule
\endlastfoot
Abnormal morphology of the nasal alae & 0.89$\pm$0.02 & 0.89$\pm$0.02 & 0.89$\pm$0.02 & 0.89$\pm$0.03 & 1210 & 0.50 \\
Abnormal oral morphology & 0.89$\pm$0.02 & 0.89$\pm$0.02 & 0.91$\pm$0.01 & 0.86$\pm$0.03 & 2176 & 0.47 \\
Abnormal cheek morphology & 0.89$\pm$0.02 & 0.89$\pm$0.02 & 0.91$\pm$0.03 & 0.86$\pm$0.04 & 712 & 0.47 \\
Abnormality of the eye & 0.89$\pm$0.02 & 0.89$\pm$0.02 & 0.91$\pm$0.02 & 0.86$\pm$0.02 & 1492 & 0.48 \\
Abnormal external nose morphology & 0.89$\pm$0.01 & 0.89$\pm$0.01 & 0.90$\pm$0.01 & 0.87$\pm$0.02 & 1722 & 0.49 \\
Abnormal eyebrow morphology & 0.89$\pm$0.03 & 0.88$\pm$0.03 & 0.91$\pm$0.04 & 0.86$\pm$0.03 & 1660 & 0.47 \\
Abnormality of the nose & 0.88$\pm$0.01 & 0.88$\pm$0.01 & 0.90$\pm$0.03 & 0.86$\pm$0.03 & 2038 & 0.48 \\
Abnormality of the mouth & 0.88$\pm$0.01 & 0.88$\pm$0.02 & 0.90$\pm$0.01 & 0.86$\pm$0.02 & 2178 & 0.47 \\
\end{longtable}

\section{Evaluation of Generalizability}\label{secA2}

\setlength{\tabcolsep}{3pt}
\begin{longtable}{rcccccccccc}
\caption{An overview of the mean F1-Score performance difference (test set vs. validation set) of the HPO models with a prevalence greater than zero. A number below zero means the performance was better on the test set, while a number above zero means the performance on the validation set was better. The labels in bold letters are leaf nodes of the Human Phenotype Ontology. The names of the unseen disorders are as follows: Susceptibility to Autism (StA), FBXW7 syndrome (FBXW7S), Floating-Harbor syndrome (FLHS), Mowat-Wilson syndrome (MOWS), Nicolaides-Baraitser syndrome (NCBRS), Ohdo syndrome, SBBYS-variant (SBBYSS), Opitz GBBB syndrome (OGBBBS), Sotos syndrome (SOTOS), Seckel syndrome (SS), and White-Sutton syndrome (WHSUS).} \label{tab:test_vs_val_unseen_appendix} \\
\toprule
HPO & StA & FBXW7S & FLHS & MOWS & NCBRS & SBBYSS & OGBBBS & SOTOS & SS & WHSUS \\
\midrule
\endfirsthead
\caption[]{An overview of the mean F1-Score performance difference (test set vs. validation set) of the HPO models with a prevalence greater than zero. A number below zero means the performance was better on the test set, while a number above zero means the performance on the validation set was better. The labels in bold letters are leaf nodes of the Human Phenotype Ontology. The names of the unseen disorders are as follows: Susceptibility to Autism (StA), FBXW7 syndrome (FBXW7S), Floating-Harbor syndrome (FLHS), Mowat-Wilson syndrome (MOWS), Nicolaides-Baraitser syndrome (NCBRS), Ohdo syndrome, SBBYS-variant (SBBYSS), Opitz GBBB syndrome (OGBBBS), Sotos syndrome (SOTOS), Seckel syndrome (SS), and White-Sutton syndrome (WHSUS).} \\
\toprule
HPO & StA & FBXW7S & FLHS & MOWS & NCBRS & SBBYSS & OGBBBS & SOTOS & SS & WHSUS \\
\midrule
\endhead
\midrule
\multicolumn{11}{r}{Continued on next page} \\
\midrule
\endfoot
\bottomrule
\endlastfoot
Abnormality of the face & \textbf{-0.12} & \textbf{-0.12} & 0.13 & -0.01 & 0.02 & \textbf{-0.12} & \textbf{-0.12} & \textbf{-0.12} & \textbf{-0.12} & 0.08 \\
Abnormal oral cavity morphology & 0.20 & 0.30 & 0.20 & 0.12 & 0.20 & 0.12 & 0.30 & N/A & \textbf{-0.13} & 0.37 \\
Abnormality of the mouth & 0.21 & 0.31 & 0.21 & 0.13 & 0.21 & 0.13 & 0.31 & N/A & \textbf{-0.12} & 0.38 \\
Abnormal oral morphology & 0.22 & 0.32 & 0.22 & \textbf{0.14} & 0.22 & \textbf{0.14} & 0.32 & N/A & 0.22 & 0.39 \\
Abnormal lip morphology & 0.21 & 0.55 & 0.02 & 0.31 & 0.21 & N/A & 0.31 & N/A & \textbf{-0.12} & 0.08 \\
Abnormal nasal morphology & N/A & -0.01 & 0.31 & 0.21 & 0.02 & -0.01 & \textbf{-0.12} & N/A & N/A & 0.38 \\
Abnormality of the nose & N/A & -0.01 & 0.55 & 0.21 & 0.02 & -0.01 & \textbf{-0.12} & N/A & N/A & 0.38 \\
Abnormality of the eye & 0.22 & 0.32 & N/A & \textbf{-0.11} & 0.89 & \textbf{-0.11} & \textbf{-0.11} & \textbf{-0.11} & N/A & 0.22 \\
\end{longtable}

\end{landscape}

\begin{longtable}{rccccccc}
\caption{An overview of the mean F1-Score performance differences (test set vs. validation set) of the HPO models with a prevalence greater than zero. A number below zero means the performance was better on the test set, while a number above zero means the performance on the validation set was better. The labels in bold letters are leaf nodes of the Human Phenotype Ontology. The names of the seen disorders are as follows: Coffin-Siris syndrome (CSS), Cornelia de Lange syndrome (CdLS), Hyperphosphatasia with mental retardation syndrome (HPMRS), KBG syndrome (KBGS), Noonan syndrome (NS), Ogden syndrome (OGDNS), and Williams-Beuren syndrome (WBS).} \label{tab:test_vs_val_seen_appendix} \\
\toprule
HPO & CSS & CdLS & HPMRS & KBGS & NS & OGDNS & WBS \\
\midrule
\endfirsthead
\caption[]{An overview of the mean F1-Score performance differences (test set vs. validation set) of the HPO models with a prevalence greater than zero. A number below zero means the performance was better on the test set, while a number above zero means the performance on the validation set was better. The labels in bold letters are leaf nodes of the Human Phenotype Ontology. The names of the seen disorders are as follows: Coffin-Siris syndrome (CSS), Cornelia de Lange syndrome (CdLS), Hyperphosphatasia with mental retardation syndrome (HPMRS), KBG syndrome (KBGS), Noonan syndrome (NS), Ogden syndrome (OGDNS), and Williams-Beuren syndrome (WBS).} \\
\toprule
HPO & CSS & CdLS & HPMRS & KBGS & NS & OGDNS & WBS \\
\midrule
\endhead
\midrule
\multicolumn{8}{r}{Continued on next page} \\
\midrule
\endfoot
\bottomrule
\endlastfoot
Abnormality of the face & -0.07 & \textbf{-0.12} & \textbf{-0.12} & 0.08 & \textbf{-0.12} & \textbf{-0.12} & \textbf{-0.12} \\
Abnormal oral cavity morphology & -0.08 & \textbf{-0.13} & -0.02 & 0.37 & 0.07 & 0.01 & \textbf{-0.13} \\
Abnormality of the mouth & -0.07 & \textbf{-0.12} & -0.01 & 0.38 & 0.21 & 0.02 & \textbf{-0.12} \\
Abnormal oral morphology & -0.06 & \textbf{-0.11} & 0.00 & 0.39 & 0.09 & 0.03 & \textbf{-0.11} \\
Abnormal lip morphology & \textbf{-0.12} & \textbf{-0.12} & 0.55 & 0.38 & 0.08 & 0.02 & \textbf{-0.12} \\
Abnormal nasal morphology & \textbf{-0.12} & 0.13 & -0.01 & 0.21 & 0.38 & 0.02 & 0.02 \\
Abnormality of the nose & \textbf{-0.12} & 0.13 & -0.01 & 0.21 & 0.38 & 0.02 & 0.02 \\
Abnormality of the eye & 0.32 & N/A & 0.32 & N/A & 0.03 & 0.22 & \textbf{-0.11} \\
\end{longtable}

\begin{longtable}{rccccccc}
\caption{Prediction results on the test set for Susceptibility to Autism. The HPO terms in bold letters are leafs of the human phenotype ontology. F1: F1-Score, Prec: Precision, Reca: Recall, N: Samples, Prev: Prevalence, Det. P: Detection Prevalence.} \label{tab:pred_result_StA} \\
\toprule
 & AUROC & F1 & Prec & Reca & N & Prev & Det. P \\
\midrule
\endfirsthead
\caption[]{Prediction results on the test set for Susceptibility to Autism. The HPO terms in bold letters are leafs of the human phenotype ontology. F1: F1-Score, Prec: Precision, Reca: Recall, N: Samples, Prev: Prevalence, Det. P: Detection Prevalence.} \\
\toprule
 & AUROC & F1 & Prec & Reca & N & Prev & Det. P \\
\midrule
\endhead
\midrule
\multicolumn{8}{r}{Continued on next page} \\
\midrule
\endfoot
\bottomrule
\endlastfoot
Abnormality of the face & N/A & 1.00 & 1.00 & 1.00 & 2 & 1.00 & 1.00 \\
Abnormality of the mouth & 0.50 & 0.67 & 0.50 & 1.00 & 2 & 0.50 & 1.00 \\
Abnormal oral morphology & 0.50 & 0.67 & 0.50 & 1.00 & 2 & 0.50 & 1.00 \\
Abnormal oral cavity morphology & 0.50 & 0.67 & 0.50 & 1.00 & 2 & 0.50 & 1.00 \\
Abnormal lip morphology & 0.50 & 0.67 & 0.50 & 1.00 & 2 & 0.50 & 1.00 \\
Abnormal lower lip morphology & 0.50 & 0.67 & 0.50 & 1.00 & 2 & 0.50 & 1.00 \\
\textbf{Thick lower lip vermilion} & 0.50 & 0.67 & 0.50 & 1.00 & 2 & 0.50 & 1.00 \\
Abnormality of the eye & 0.50 & 0.67 & 0.50 & 1.00 & 2 & 0.50 & 1.00 \\
Abnormal eye morphology & 0.50 & 0.67 & 0.50 & 1.00 & 2 & 0.50 & 1.00 \\
Abnormality of globe location & 0.50 & 0.67 & 0.50 & 1.00 & 2 & 0.50 & 1.00 \\
\textbf{Hypertelorism} & 0.50 & 0.67 & 0.50 & 1.00 & 2 & 0.50 & 1.00 \\
\end{longtable}

\begin{longtable}{rccccccc}
\caption{Prediction results on the test set for FBXW7 syndrome. The HPO terms in bold letters are leafs of the human phenotype ontology. F1: F1-Score, Prec: Precision, Reca: Recall, N: Samples, Prev: Prevalence, Det. P: Detection Prevalence.} \label{tab:pred_result_FBXW7S} \\
\toprule
 & AUROC & F1 & Prec & Reca & N & Prev & Det. P \\
\midrule
\endfirsthead
\caption[]{Prediction results on the test set for FBXW7 syndrome. The HPO terms in bold letters are leafs of the human phenotype ontology. F1: F1-Score, Prec: Precision, Reca: Recall, N: Samples, Prev: Prevalence, Det. P: Detection Prevalence.} \\
\toprule
 & AUROC & F1 & Prec & Reca & N & Prev & Det. P \\
\midrule
\endhead
\midrule
\multicolumn{8}{r}{Continued on next page} \\
\midrule
\endfoot
\bottomrule
\endlastfoot
Abnormality of the face & N/A & 1.00 & 1.00 & 1.00 & 5 & 1.00 & 1.00 \\
Abnormality of the mouth & 0.50 & 0.57 & 0.40 & 1.00 & 5 & 0.40 & 1.00 \\
Abnormal oral morphology & 0.50 & 0.57 & 0.40 & 1.00 & 5 & 0.40 & 1.00 \\
Abnormal oral cavity morphology & 0.50 & 0.57 & 0.40 & 1.00 & 5 & 0.40 & 1.00 \\
Abnormal lip morphology & 0.50 & 0.33 & 0.20 & 1.00 & 5 & 0.20 & 1.00 \\
Abnormal lower lip morphology & 0.50 & 0.33 & 0.20 & 1.00 & 5 & 0.20 & 1.00 \\
\textbf{Thick lower lip vermilion} & 0.50 & 0.33 & 0.20 & 1.00 & 5 & 0.20 & 1.00 \\
Thin vermilion border & 0.12 & 0.00 & 0.00 & 0.00 & 5 & 0.20 & 0.60 \\
\end{longtable}

\begin{longtable}{rccccccc}
\caption{Prediction results on the test set for Floating-Harbor syndrome. The HPO terms in bold letters are leafs of the human phenotype ontology. F1: F1-Score, Prec: Precision, Reca: Recall, N: Samples, Prev: Prevalence, Det. P: Detection Prevalence.} \label{tab:pred_result_FLHS} \\
\toprule
 & AUROC & F1 & Prec & Reca & N & Prev & Det. P \\
\midrule
\endfirsthead
\caption[]{Prediction results on the test set for Floating-Harbor syndrome. The HPO terms in bold letters are leafs of the human phenotype ontology. F1: F1-Score, Prec: Precision, Reca: Recall, N: Samples, Prev: Prevalence, Det. P: Detection Prevalence.} \\
\toprule
 & AUROC & F1 & Prec & Reca & N & Prev & Det. P \\
\midrule
\endhead
\midrule
\multicolumn{8}{r}{Continued on next page} \\
\midrule
\endfoot
\bottomrule
\endlastfoot
Abnormality of the face & N/A & 0.75 & 1.00 & 0.60 & 5 & 1.00 & 0.60 \\
Abnormality of the mouth & 0.75 & 0.67 & 1.00 & 0.50 & 5 & 0.80 & 0.40 \\
Abnormal oral morphology & 0.75 & 0.67 & 1.00 & 0.50 & 5 & 0.80 & 0.40 \\
Abnormal oral cavity morphology & 0.75 & 0.67 & 1.00 & 0.50 & 5 & 0.80 & 0.40 \\
Abnormal lip morphology & 0.75 & 0.86 & 0.75 & 1.00 & 5 & 0.60 & 0.80 \\
Abnormal upper lip morphology & 0.83 & 0.80 & 1.00 & 0.67 & 5 & 0.60 & 0.40 \\
Abnormality of the philtrum & 0.83 & 0.80 & 1.00 & 0.67 & 5 & 0.60 & 0.40 \\
\textbf{Short philtrum} & 0.42 & 0.40 & 0.50 & 0.33 & 5 & 0.60 & 0.40 \\
\end{longtable}

\begin{longtable}{rccccccc}
\caption{Prediction results on the test set for Mowat-Wilson syndrome. The HPO terms in bold letters are leafs of the human phenotype ontology. F1: F1-Score, Prec: Precision, Reca: Recall, N: Samples, Prev: Prevalence, Det. P: Detection Prevalence.} \label{tab:pred_result_MOWS} \\
\toprule
 & AUROC & F1 & Prec & Reca & N & Prev & Det. P \\
\midrule
\endfirsthead
\caption[]{Prediction results on the test set for Mowat-Wilson syndrome. The HPO terms in bold letters are leafs of the human phenotype ontology. F1: F1-Score, Prec: Precision, Reca: Recall, N: Samples, Prev: Prevalence, Det. P: Detection Prevalence.} \\
\toprule
 & AUROC & F1 & Prec & Reca & N & Prev & Det. P \\
\midrule
\endhead
\midrule
\multicolumn{8}{r}{Continued on next page} \\
\midrule
\endfoot
\bottomrule
\endlastfoot
Abnormality of the face & N/A & 0.89 & 1.00 & 0.80 & 5 & 1.00 & 0.80 \\
Abnormality of the mouth & 0.38 & 0.75 & 0.75 & 0.75 & 5 & 0.80 & 0.80 \\
Abnormal oral morphology & 0.38 & 0.75 & 0.75 & 0.75 & 5 & 0.80 & 0.80 \\
Abnormal oral cavity morphology & 0.38 & 0.75 & 0.75 & 0.75 & 5 & 0.80 & 0.80 \\
Abnormal lip morphology & 0.33 & 0.57 & 0.50 & 0.67 & 5 & 0.60 & 0.80 \\
Abnormal lower lip morphology & 0.33 & 0.57 & 0.50 & 0.67 & 5 & 0.60 & 0.80 \\
\textbf{Thick lower lip vermilion} & 0.33 & 0.57 & 0.50 & 0.67 & 5 & 0.60 & 0.80 \\
\textbf{Everted lower lip vermilion} & 0.25 & 0.33 & 0.25 & 0.50 & 5 & 0.40 & 0.80 \\
\end{longtable}

\begin{longtable}{rccccccc}
\caption{Prediction results on the test set for Nicolaides-Baraitser syndrome. The HPO terms in bold letters are leafs of the human phenotype ontology. F1: F1-Score, Prec: Precision, Reca: Recall, N: Samples, Prev: Prevalence, Det. P: Detection Prevalence.} \label{tab:pred_result_NCBRS} \\
\toprule
 & AUROC & F1 & Prec & Reca & N & Prev & Det. P \\
\midrule
\endfirsthead
\caption[]{Prediction results on the test set for Nicolaides-Baraitser syndrome. The HPO terms in bold letters are leafs of the human phenotype ontology. F1: F1-Score, Prec: Precision, Reca: Recall, N: Samples, Prev: Prevalence, Det. P: Detection Prevalence.} \\
\toprule
 & AUROC & F1 & Prec & Reca & N & Prev & Det. P \\
\midrule
\endhead
\midrule
\multicolumn{8}{r}{Continued on next page} \\
\midrule
\endfoot
\bottomrule
\endlastfoot
Abnormality of the face & N/A & 0.86 & 1.00 & 0.75 & 4 & 1.00 & 0.75 \\
Abnormality of the mouth & 0.33 & 0.67 & 0.67 & 0.67 & 4 & 0.75 & 0.75 \\
Abnormal oral morphology & 0.33 & 0.67 & 0.67 & 0.67 & 4 & 0.75 & 0.75 \\
Abnormal oral cavity morphology & 0.33 & 0.67 & 0.67 & 0.67 & 4 & 0.75 & 0.75 \\
Abnormal lip morphology & 0.33 & 0.67 & 0.67 & 0.67 & 4 & 0.75 & 0.75 \\
Abnormal upper lip morphology & 0.67 & 0.50 & 0.33 & 1.00 & 4 & 0.25 & 0.75 \\
Abnormality of the philtrum & 0.67 & 0.50 & 0.33 & 1.00 & 4 & 0.25 & 0.75 \\
\textbf{Long philtrum} & 0.67 & 0.50 & 0.33 & 1.00 & 4 & 0.25 & 0.75 \\
\end{longtable}

\begin{longtable}{rccccccc}
\caption{Prediction results on the test set for Ohdo syndrome, SBBYS-variant. The HPO terms in bold letters are leafs of the human phenotype ontology. F1: F1-Score, Prec: Precision, Reca: Recall, N: Samples, Prev: Prevalence, Det. P: Detection Prevalence.} \label{tab:pred_result_SBBYSS} \\
\toprule
 & AUROC & F1 & Prec & Reca & N & Prev & Det. P \\
\midrule
\endfirsthead
\caption[]{Prediction results on the test set for Ohdo syndrome, SBBYS-variant. The HPO terms in bold letters are leafs of the human phenotype ontology. F1: F1-Score, Prec: Precision, Reca: Recall, N: Samples, Prev: Prevalence, Det. P: Detection Prevalence.} \\
\toprule
 & AUROC & F1 & Prec & Reca & N & Prev & Det. P \\
\midrule
\endhead
\midrule
\multicolumn{8}{r}{Continued on next page} \\
\midrule
\endfoot
\bottomrule
\endlastfoot
Abnormality of the face & N/A & 1.00 & 1.00 & 1.00 & 5 & 1.00 & 1.00 \\
Abnormality of the mouth & 0.50 & 0.75 & 0.60 & 1.00 & 5 & 0.60 & 1.00 \\
Abnormal oral morphology & 0.50 & 0.75 & 0.60 & 1.00 & 5 & 0.60 & 1.00 \\
Abnormal oral cavity morphology & 0.50 & 0.75 & 0.60 & 1.00 & 5 & 0.60 & 1.00 \\
Abnormality of mouth size & 0.50 & 0.75 & 0.60 & 1.00 & 5 & 0.60 & 1.00 \\
\textbf{Wide mouth} & 0.50 & 0.57 & 0.40 & 1.00 & 5 & 0.40 & 1.00 \\
\textbf{Narrow mouth} & 0.00 & 0.00 & 0.00 & 0.00 & 5 & 0.20 & 0.80 \\
Abnormality of the orbital region & 0.50 & 0.57 & 0.40 & 1.00 & 5 & 0.40 & 1.00 \\
\end{longtable}

\begin{longtable}{rccccccc}
\caption{Prediction results on the test set for Opitz GBBB syndrome. The HPO terms in bold letters are leafs of the human phenotype ontology. F1: F1-Score, Prec: Precision, Reca: Recall, N: Samples, Prev: Prevalence, Det. P: Detection Prevalence.} \label{tab:pred_result_OGBBBS} \\
\toprule
 & AUROC & F1 & Prec & Reca & N & Prev & Det. P \\
\midrule
\endfirsthead
\caption[]{Prediction results on the test set for Opitz GBBB syndrome. The HPO terms in bold letters are leafs of the human phenotype ontology. F1: F1-Score, Prec: Precision, Reca: Recall, N: Samples, Prev: Prevalence, Det. P: Detection Prevalence.} \\
\toprule
 & AUROC & F1 & Prec & Reca & N & Prev & Det. P \\
\midrule
\endhead
\midrule
\multicolumn{8}{r}{Continued on next page} \\
\midrule
\endfoot
\bottomrule
\endlastfoot
Abnormality of the face & N/A & 1.00 & 1.00 & 1.00 & 5 & 1.00 & 1.00 \\
Abnormality of the mouth & 0.50 & 0.57 & 0.40 & 1.00 & 5 & 0.40 & 1.00 \\
Abnormal oral morphology & 0.50 & 0.57 & 0.40 & 1.00 & 5 & 0.40 & 1.00 \\
Abnormal oral cavity morphology & 0.50 & 0.57 & 0.40 & 1.00 & 5 & 0.40 & 1.00 \\
Abnormal lip morphology & 0.50 & 0.57 & 0.40 & 1.00 & 5 & 0.40 & 1.00 \\
Abnormal upper lip morphology & 0.50 & 0.57 & 0.40 & 1.00 & 5 & 0.40 & 1.00 \\
Abnormality of the philtrum & 0.50 & 0.57 & 0.40 & 1.00 & 5 & 0.40 & 1.00 \\
\textbf{Long philtrum} & 0.25 & 0.33 & 0.25 & 0.50 & 5 & 0.40 & 0.80 \\
\end{longtable}

\begin{longtable}{rccccccc}
\caption{Prediction results on the test set for Sotos syndrome. The HPO terms in bold letters are leafs of the human phenotype ontology. F1: F1-Score, Prec: Precision, Reca: Recall, N: Samples, Prev: Prevalence, Det. P: Detection Prevalence.} \label{tab:pred_result_SOTOS} \\
\toprule
 & AUROC & F1 & Prec & Reca & N & Prev & Det. P \\
\midrule
\endfirsthead
\caption[]{Prediction results on the test set for Sotos syndrome. The HPO terms in bold letters are leafs of the human phenotype ontology. F1: F1-Score, Prec: Precision, Reca: Recall, N: Samples, Prev: Prevalence, Det. P: Detection Prevalence.} \\
\toprule
 & AUROC & F1 & Prec & Reca & N & Prev & Det. P \\
\midrule
\endhead
\midrule
\multicolumn{8}{r}{Continued on next page} \\
\midrule
\endfoot
\bottomrule
\endlastfoot
Abnormality of the face & N/A & 1.00 & 1.00 & 1.00 & 4 & 1.00 & 1.00 \\
Abnormality of the orbital region & N/A & 1.00 & 1.00 & 1.00 & 4 & 1.00 & 1.00 \\
Abnormality of the ocular adnexa & N/A & 1.00 & 1.00 & 1.00 & 4 & 1.00 & 1.00 \\
Abnormal ocular adnexa morphology & N/A & 0.86 & 1.00 & 0.75 & 4 & 1.00 & 0.75 \\
Abnormal eyelid morphology & N/A & 0.86 & 1.00 & 0.75 & 4 & 1.00 & 0.75 \\
\textbf{Epicanthus} & 0.67 & 0.50 & 0.33 & 1.00 & 4 & 0.25 & 0.75 \\
Abnormality of the palpebral fissures & N/A & 1.00 & 1.00 & 1.00 & 4 & 1.00 & 1.00 \\
Slanting of the palpebral fissure & N/A & 1.00 & 1.00 & 1.00 & 4 & 1.00 & 1.00 \\
\end{longtable}

\begin{longtable}{rccccccc}
\caption{Prediction results on the test set for Seckel syndrome. The HPO terms in bold letters are leafs of the human phenotype ontology. F1: F1-Score, Prec: Precision, Reca: Recall, N: Samples, Prev: Prevalence, Det. P: Detection Prevalence.} \label{tab:pred_result_SS} \\
\toprule
 & AUROC & F1 & Prec & Reca & N & Prev & Det. P \\
\midrule
\endfirsthead
\caption[]{Prediction results on the test set for Seckel syndrome. The HPO terms in bold letters are leafs of the human phenotype ontology. F1: F1-Score, Prec: Precision, Reca: Recall, N: Samples, Prev: Prevalence, Det. P: Detection Prevalence.} \\
\toprule
 & AUROC & F1 & Prec & Reca & N & Prev & Det. P \\
\midrule
\endhead
\midrule
\multicolumn{8}{r}{Continued on next page} \\
\midrule
\endfoot
\bottomrule
\endlastfoot
Abnormality of the face & N/A & 1.00 & 1.00 & 1.00 & 2 & 1.00 & 1.00 \\
Abnormality of the mouth & N/A & 1.00 & 1.00 & 1.00 & 2 & 1.00 & 1.00 \\
Abnormal oral morphology & N/A & 0.67 & 1.00 & 0.50 & 2 & 1.00 & 0.50 \\
Abnormal oral cavity morphology & N/A & 1.00 & 1.00 & 1.00 & 2 & 1.00 & 1.00 \\
Abnormal lip morphology & N/A & 1.00 & 1.00 & 1.00 & 2 & 1.00 & 1.00 \\
Abnormal upper lip morphology & N/A & 1.00 & 1.00 & 1.00 & 2 & 1.00 & 1.00 \\
Abnormality of the philtrum & N/A & 0.67 & 1.00 & 0.50 & 2 & 1.00 & 0.50 \\
\textbf{Long philtrum} & N/A & 1.00 & 1.00 & 1.00 & 2 & 1.00 & 1.00 \\
Thin vermilion border & 0.50 & 0.67 & 0.50 & 1.00 & 2 & 0.50 & 1.00 \\
\textbf{Thin upper lip vermilion} & 0.50 & 0.67 & 0.50 & 1.00 & 2 & 0.50 & 1.00 \\
Abnormal facial shape & N/A & 1.00 & 1.00 & 1.00 & 2 & 1.00 & 1.00 \\
Small face & N/A & 0.67 & 1.00 & 0.50 & 2 & 1.00 & 0.50 \\
\textbf{Narrow face} & N/A & 1.00 & 1.00 & 1.00 & 2 & 1.00 & 1.00 \\
\end{longtable}

\begin{longtable}{rccccccc}
\caption{Prediction results on the test set for White-Sutton syndrome. The HPO terms in bold letters are leafs of the human phenotype ontology. F1: F1-Score, Prec: Precision, Reca: Recall, N: Samples, Prev: Prevalence, Det. P: Detection Prevalence.} \label{tab:pred_result_WHSUS} \\
\toprule
 & AUROC & F1 & Prec & Reca & N & Prev & Det. P \\
\midrule
\endfirsthead
\caption[]{Prediction results on the test set for White-Sutton syndrome. The HPO terms in bold letters are leafs of the human phenotype ontology. F1: F1-Score, Prec: Precision, Reca: Recall, N: Samples, Prev: Prevalence, Det. P: Detection Prevalence.} \\
\toprule
 & AUROC & F1 & Prec & Reca & N & Prev & Det. P \\
\midrule
\endhead
\midrule
\multicolumn{8}{r}{Continued on next page} \\
\midrule
\endfoot
\bottomrule
\endlastfoot
Abnormality of the face & N/A & 0.80 & 1.00 & 0.67 & 3 & 1.00 & 0.67 \\
Abnormality of the mouth & N/A & 0.50 & 1.00 & 0.33 & 3 & 1.00 & 0.33 \\
Abnormal oral morphology & N/A & 0.50 & 1.00 & 0.33 & 3 & 1.00 & 0.33 \\
Abnormal oral cavity morphology & N/A & 0.50 & 1.00 & 0.33 & 3 & 1.00 & 0.33 \\
Abnormal lip morphology & N/A & 0.80 & 1.00 & 0.67 & 3 & 1.00 & 0.67 \\
Abnormal lower lip morphology & 0.75 & 0.67 & 1.00 & 0.50 & 3 & 0.67 & 0.33 \\
\textbf{Thick lower lip vermilion} & 0.50 & 0.50 & 0.33 & 1.00 & 3 & 0.33 & 1.00 \\
\textbf{Everted lower lip vermilion} & 1.00 & 1.00 & 1.00 & 1.00 & 3 & 0.67 & 0.67 \\
\end{longtable}

\begin{longtable}{rccccccc}
\caption{Prediction results on the test set for Coffin-Siris syndrome. The HPO terms in bold letters are leafs of the human phenotype ontology. F1: F1-Score, Prec: Precision, Reca: Recall, N: Samples, Prev: Prevalence, Det. P: Detection Prevalence.} \label{tab:pred_result_CSS} \\
\toprule
 & AUROC & F1 & Prec & Reca & N & Prev & Det. P \\
\midrule
\endfirsthead
\caption[]{Prediction results on the test set for Coffin-Siris syndrome. The HPO terms in bold letters are leafs of the human phenotype ontology. F1: F1-Score, Prec: Precision, Reca: Recall, N: Samples, Prev: Prevalence, Det. P: Detection Prevalence.} \\
\toprule
 & AUROC & F1 & Prec & Reca & N & Prev & Det. P \\
\midrule
\endhead
\midrule
\multicolumn{8}{r}{Continued on next page} \\
\midrule
\endfoot
\bottomrule
\endlastfoot
Abnormality of the face & N/A & 0.95 & 1.00 & 0.90 & 10 & 1.00 & 0.90 \\
Abnormality of the mouth & N/A & 0.95 & 1.00 & 0.90 & 10 & 1.00 & 0.90 \\
Abnormal oral morphology & N/A & 0.95 & 1.00 & 0.90 & 10 & 1.00 & 0.90 \\
Abnormal oral cavity morphology & N/A & 0.95 & 1.00 & 0.90 & 10 & 1.00 & 0.90 \\
Abnormal lip morphology & N/A & 1.00 & 1.00 & 1.00 & 10 & 1.00 & 1.00 \\
Abnormal upper lip morphology & 1.00 & 1.00 & 1.00 & 1.00 & 10 & 0.90 & 0.90 \\
Abnormality of the philtrum & 1.00 & 1.00 & 1.00 & 1.00 & 10 & 0.90 & 0.90 \\
\textbf{Broad philtrum} & 1.00 & 1.00 & 1.00 & 1.00 & 10 & 0.90 & 0.90 \\
\end{longtable}

\begin{longtable}{rccccccc}
\caption{Prediction results on the test set for Cornelia de Lange syndrome. The HPO terms in bold letters are leafs of the human phenotype ontology. F1: F1-Score, Prec: Precision, Reca: Recall, N: Samples, Prev: Prevalence, Det. P: Detection Prevalence.} \label{tab:pred_result_CdLS} \\
\toprule
 & AUROC & F1 & Prec & Reca & N & Prev & Det. P \\
\midrule
\endfirsthead
\caption[]{Prediction results on the test set for Cornelia de Lange syndrome. The HPO terms in bold letters are leafs of the human phenotype ontology. F1: F1-Score, Prec: Precision, Reca: Recall, N: Samples, Prev: Prevalence, Det. P: Detection Prevalence.} \\
\toprule
 & AUROC & F1 & Prec & Reca & N & Prev & Det. P \\
\midrule
\endhead
\midrule
\multicolumn{8}{r}{Continued on next page} \\
\midrule
\endfoot
\bottomrule
\endlastfoot
Abnormality of the face & N/A & 1.00 & 1.00 & 1.00 & 5 & 1.00 & 1.00 \\
Abnormality of the mouth & N/A & 1.00 & 1.00 & 1.00 & 5 & 1.00 & 1.00 \\
Abnormal oral morphology & N/A & 1.00 & 1.00 & 1.00 & 5 & 1.00 & 1.00 \\
Abnormal oral cavity morphology & N/A & 1.00 & 1.00 & 1.00 & 5 & 1.00 & 1.00 \\
Abnormal lip morphology & N/A & 1.00 & 1.00 & 1.00 & 5 & 1.00 & 1.00 \\
Abnormal upper lip morphology & 0.50 & 0.75 & 0.60 & 1.00 & 5 & 0.60 & 1.00 \\
Abnormality of the philtrum & 0.50 & 0.75 & 0.60 & 1.00 & 5 & 0.60 & 1.00 \\
\textbf{Long philtrum} & 0.33 & 0.57 & 0.50 & 0.67 & 5 & 0.60 & 0.80 \\
\end{longtable}

\begin{longtable}{rccccccc}
\caption{Prediction results on the test set for Hyperphosphatasia with mental retardation syndrome. The HPO terms in bold letters are leafs of the human phenotype ontology. F1: F1-Score, Prec: Precision, Reca: Recall, N: Samples, Prev: Prevalence, Det. P: Detection Prevalence.} \label{tab:pred_result_HPMRS} \\
\toprule
 & AUROC & F1 & Prec & Reca & N & Prev & Det. P \\
\midrule
\endfirsthead
\caption[]{Prediction results on the test set for Hyperphosphatasia with mental retardation syndrome. The HPO terms in bold letters are leafs of the human phenotype ontology. F1: F1-Score, Prec: Precision, Reca: Recall, N: Samples, Prev: Prevalence, Det. P: Detection Prevalence.} \\
\toprule
 & AUROC & F1 & Prec & Reca & N & Prev & Det. P \\
\midrule
\endhead
\midrule
\multicolumn{8}{r}{Continued on next page} \\
\midrule
\endfoot
\bottomrule
\endlastfoot
Abnormality of the face & N/A & 1.00 & 1.00 & 1.00 & 5 & 1.00 & 1.00 \\
Abnormality of the mouth & 0.50 & 0.89 & 0.80 & 1.00 & 5 & 0.80 & 1.00 \\
Abnormal oral morphology & 0.50 & 0.89 & 0.80 & 1.00 & 5 & 0.80 & 1.00 \\
Abnormal oral cavity morphology & 0.50 & 0.89 & 0.80 & 1.00 & 5 & 0.80 & 1.00 \\
Abnormal lip morphology & 0.50 & 0.33 & 0.20 & 1.00 & 5 & 0.20 & 1.00 \\
Thin vermilion border & 0.50 & 0.33 & 0.20 & 1.00 & 5 & 0.20 & 1.00 \\
Abnormality of mouth shape & 0.50 & 0.75 & 0.60 & 1.00 & 5 & 0.60 & 1.00 \\
\textbf{Downturned corners of mouth} & 0.50 & 0.75 & 0.60 & 1.00 & 5 & 0.60 & 1.00 \\
\end{longtable}

\begin{longtable}{rccccccc}
\caption{Prediction results on the test set for KBG syndrome. The HPO terms in bold letters are leafs of the human phenotype ontology. F1: F1-Score, Prec: Precision, Reca: Recall, N: Samples, Prev: Prevalence, Det. P: Detection Prevalence.} \label{tab:pred_result_KBGS} \\
\toprule
 & AUROC & F1 & Prec & Reca & N & Prev & Det. P \\
\midrule
\endfirsthead
\caption[]{Prediction results on the test set for KBG syndrome. The HPO terms in bold letters are leafs of the human phenotype ontology. F1: F1-Score, Prec: Precision, Reca: Recall, N: Samples, Prev: Prevalence, Det. P: Detection Prevalence.} \\
\toprule
 & AUROC & F1 & Prec & Reca & N & Prev & Det. P \\
\midrule
\endhead
\midrule
\multicolumn{8}{r}{Continued on next page} \\
\midrule
\endfoot
\bottomrule
\endlastfoot
Abnormality of the face & N/A & 0.80 & 1.00 & 0.67 & 3 & 1.00 & 0.67 \\
Abnormality of the mouth & 0.25 & 0.50 & 0.50 & 0.50 & 3 & 0.67 & 0.67 \\
Abnormal oral morphology & 0.25 & 0.50 & 0.50 & 0.50 & 3 & 0.67 & 0.67 \\
Abnormal oral cavity morphology & 0.25 & 0.50 & 0.50 & 0.50 & 3 & 0.67 & 0.67 \\
Abnormal lip morphology & 0.25 & 0.50 & 0.50 & 0.50 & 3 & 0.67 & 0.67 \\
Thin vermilion border & 0.25 & 0.50 & 0.50 & 0.50 & 3 & 0.67 & 0.67 \\
\textbf{Thin upper lip vermilion} & 0.25 & 0.50 & 0.50 & 0.50 & 3 & 0.67 & 0.67 \\
Abnormality of the orbital region & 1.00 & 1.00 & 1.00 & 1.00 & 3 & 0.67 & 0.67 \\
\end{longtable}

\begin{longtable}{rccccccc}
\caption{Prediction results on the test set for Noonan syndrome. The HPO terms in bold letters are leafs of the human phenotype ontology. F1: F1-Score, Prec: Precision, Reca: Recall, N: Samples, Prev: Prevalence, Det. P: Detection Prevalence.} \label{tab:pred_result_NS} \\
\toprule
 & AUROC & F1 & Prec & Reca & N & Prev & Det. P \\
\midrule
\endfirsthead
\caption[]{Prediction results on the test set for Noonan syndrome. The HPO terms in bold letters are leafs of the human phenotype ontology. F1: F1-Score, Prec: Precision, Reca: Recall, N: Samples, Prev: Prevalence, Det. P: Detection Prevalence.} \\
\toprule
 & AUROC & F1 & Prec & Reca & N & Prev & Det. P \\
\midrule
\endhead
\midrule
\multicolumn{8}{r}{Continued on next page} \\
\midrule
\endfoot
\bottomrule
\endlastfoot
Abnormality of the face & N/A & 1.00 & 1.00 & 1.00 & 4 & 1.00 & 1.00 \\
Abnormality of the mouth & 0.50 & 0.67 & 0.50 & 1.00 & 4 & 0.50 & 1.00 \\
Abnormal oral morphology & 0.75 & 0.80 & 0.67 & 1.00 & 4 & 0.50 & 0.75 \\
Abnormal oral cavity morphology & 0.75 & 0.80 & 0.67 & 1.00 & 4 & 0.50 & 0.75 \\
Abnormal lip morphology & 0.75 & 0.80 & 0.67 & 1.00 & 4 & 0.50 & 0.75 \\
Abnormal upper lip morphology & 0.67 & 0.50 & 0.33 & 1.00 & 4 & 0.25 & 0.75 \\
Abnormality of the philtrum & 0.67 & 0.50 & 0.33 & 1.00 & 4 & 0.25 & 0.75 \\
\textbf{Short philtrum} & 0.67 & 0.50 & 0.33 & 1.00 & 4 & 0.25 & 0.75 \\
\end{longtable}

\begin{longtable}{rccccccc}
\caption{Prediction results on the test set for Ogden syndrome. The HPO terms in bold letters are leafs of the human phenotype ontology. F1: F1-Score, Prec: Precision, Reca: Recall, N: Samples, Prev: Prevalence, Det. P: Detection Prevalence.} \label{tab:pred_result_OGDNS} \\
\toprule
 & AUROC & F1 & Prec & Reca & N & Prev & Det. P \\
\midrule
\endfirsthead
\caption[]{Prediction results on the test set for Ogden syndrome. The HPO terms in bold letters are leafs of the human phenotype ontology. F1: F1-Score, Prec: Precision, Reca: Recall, N: Samples, Prev: Prevalence, Det. P: Detection Prevalence.} \\
\toprule
 & AUROC & F1 & Prec & Reca & N & Prev & Det. P \\
\midrule
\endhead
\midrule
\multicolumn{8}{r}{Continued on next page} \\
\midrule
\endfoot
\bottomrule
\endlastfoot
Abnormality of the face & N/A & 1.00 & 1.00 & 1.00 & 4 & 1.00 & 1.00 \\
Abnormality of the mouth & 0.50 & 0.86 & 0.75 & 1.00 & 4 & 0.75 & 1.00 \\
Abnormal oral morphology & 0.50 & 0.86 & 0.75 & 1.00 & 4 & 0.75 & 1.00 \\
Abnormal oral cavity morphology & 0.50 & 0.86 & 0.75 & 1.00 & 4 & 0.75 & 1.00 \\
Abnormal lip morphology & 0.50 & 0.86 & 0.75 & 1.00 & 4 & 0.75 & 1.00 \\
Abnormal upper lip morphology & 0.50 & 0.86 & 0.75 & 1.00 & 4 & 0.75 & 1.00 \\
Abnormality of the philtrum & 0.50 & 0.40 & 0.25 & 1.00 & 4 & 0.25 & 1.00 \\
\textbf{Long philtrum} & 0.50 & 0.40 & 0.25 & 1.00 & 4 & 0.25 & 1.00 \\
\end{longtable}

\begin{longtable}{rccccccc}
\caption{Prediction results on the test set for Williams-Beuren syndrome. The HPO terms in bold letters are leafs of the human phenotype ontology. F1: F1-Score, Prec: Precision, Reca: Recall, N: Samples, Prev: Prevalence, Det. P: Detection Prevalence.} \label{tab:pred_result_WBS} \\
\toprule
 & AUROC & F1 & Prec & Reca & N & Prev & Det. P \\
\midrule
\endfirsthead
\caption[]{Prediction results on the test set for Williams-Beuren syndrome. The HPO terms in bold letters are leafs of the human phenotype ontology. F1: F1-Score, Prec: Precision, Reca: Recall, N: Samples, Prev: Prevalence, Det. P: Detection Prevalence.} \\
\toprule
 & AUROC & F1 & Prec & Reca & N & Prev & Det. P \\
\midrule
\endhead
\midrule
\multicolumn{8}{r}{Continued on next page} \\
\midrule
\endfoot
\bottomrule
\endlastfoot
Abnormality of the face & N/A & 1.00 & 1.00 & 1.00 & 4 & 1.00 & 1.00 \\
Abnormality of the mouth & N/A & 1.00 & 1.00 & 1.00 & 4 & 1.00 & 1.00 \\
Abnormal oral morphology & N/A & 1.00 & 1.00 & 1.00 & 4 & 1.00 & 1.00 \\
Abnormal oral cavity morphology & N/A & 1.00 & 1.00 & 1.00 & 4 & 1.00 & 1.00 \\
Abnormal lip morphology & N/A & 1.00 & 1.00 & 1.00 & 4 & 1.00 & 1.00 \\
Abnormal upper lip morphology & 0.50 & 0.86 & 0.75 & 1.00 & 4 & 0.75 & 1.00 \\
Abnormality of the philtrum & 0.50 & 0.86 & 0.75 & 1.00 & 4 & 0.75 & 1.00 \\
\textbf{Long philtrum} & 0.50 & 0.86 & 0.75 & 1.00 & 4 & 0.75 & 1.00 \\
\end{longtable}




\end{appendices}


\bibliography{sn-bibliography}

\end{document}